\documentclass[conference]{IEEEtran}
\IEEEoverridecommandlockouts
\usepackage{cite}
\usepackage{amsmath,amssymb,amsfonts,amsthm}
\usepackage{algorithmic}
\usepackage{graphicx}
\usepackage{textcomp}
\usepackage{xcolor}
\usepackage{balance}
\usepackage{kotex}
\usepackage{todonotes}
\usepackage{subcaption}
\usepackage{url}
\usepackage{booktabs}
\usepackage{multirow}
\usepackage{pgfplots}
\usepackage{nicefrac}
\usepackage[algo2e,ruled,vlined]{algorithm2e}
\usepackage{algorithm}
\usepackage{dsfont}
\usepackage{framed}
\usepackage{thmtools}
\usepackage{color, colortbl}

\newcommand{\ours}{\textsf{ADAPT}}

\def\BibTeX{{\rm B\kern-.05em{\sc i\kern-.025em b}\kern-.08em
    T\kern-.1667em\lower.7ex\hbox{E}\kern-.125emX}}
\begin{document}

\title{Customs Fraud Detection in the Presence of Concept Drift\\}


\author{%
  \IEEEauthorblockN{%
    Tung-Duong Mai\textsuperscript{*},
    Kien Hoang\textsuperscript{*},
    Aitolkyn Baigutanova\textsuperscript{*},
    Gaukhartas Alina\textsuperscript{*}, 
  }%
  \IEEEauthorblockA{\textit{Korea Advanced Institute of Science and Technology} \\ 
  \{maitungduong2605, kienapp286, baiaitolkyn, gaukhar112200\}@gmail.com}%
  \and
  \IEEEauthorblockN{%
    Sundong Kim %
  }%
  \IEEEauthorblockA{\textit{Institute for Basic Science} \\ sundong@ibs.re.kr}%
}


\maketitle
\begingroup\renewcommand\thefootnote{*}
\footnotetext{Work done while interning at Institute for Basic Science}
\endgroup

\begin{abstract}
Capturing the changing trade pattern is critical in customs fraud detection. As new goods are imported and novel frauds arise, a drift-aware fraud detection system is needed to detect both known frauds and unknown frauds within a limited budget. The current paper proposes \ours{}, an adaptive selection method that controls the balance between exploitation and exploration strategies used for customs fraud detection. \ours{} makes use of the model performance trends and the amount of concept drift to determine the best exploration ratio at every time. Experiments on data from four countries over several years show that each country requires a different amount of exploration for maintaining its fraud detection system. We find the system with \ours{} can gradually adapt to the dataset and find the appropriate amount of exploration ratio with high performance.
\end{abstract} 

\begin{IEEEkeywords}
Concept Drift, Customs Fraud Detection, Exploration-Exploitation Dilemma, Multi-Armed Bandit
\end{IEEEkeywords}
\section{Introduction}


With enormous daily trade traffic, effective trade handling becomes the main task of customs administrations, and the urge for AI-based fraud detection mechanisms becomes apparent~\cite{sisam2015paper, mikuriya2021wcj}. However, maintaining a sustainable fraud detection system is challenging due to the changes in customs trades and fraudulent transaction trends with time. A fraud detection model trained on historical data often falls into a confirmation bias and results in performance degradation~\cite{kim2020take}. The change in data characteristics and distribution over time is referred to as concept drift, which can be gradual, incremental, recurrent, or sudden~\cite{lu2019conceptdrift}. In customs, concept drift is caused by alterations of importers, goods types, and business partners~\cite{changing2011imf}.

The customs workflow follows the human-in-the-loop inspection format, where physical inspections are carried out with the help of a fraud detection model, and customs officers confirm whether the declared item is fraud or not. If the inspection reveals fraud, the officers can levy extra duties and the results will be used to update the fraud detection model. Usually, the model provides the most suspicious items for inspection, which can be problematic in countries facing concept drift in their trade pattern.

While the fraud detection model should keep on catching the known frauds and secure the revenue, it should also acquire knowledge about new fraudulent behavior. These conflicting objectives can be described as the famous exploration-exploitation dilemma~\cite{AUDIBERT20091876}, where there must be a balance between selecting illicit items to secure immediate revenue (exploitation) and discovering new items to maintain long-term performance (exploration). 

One practical solution is to randomly inspect a small set of declared items. This is called random selection, which can estimate their trade statistics and learn new fraud patterns~\cite{han2014kcs}. Adding some random exploration together with exploitation strategy is shown to be effective in the presence of concept drift~\cite{kim2020take}. However, there are no studies on how to determine the best exploration amount in customs settings. Empirically setting the exploration ratio is problematic where the amount of concept drift in the underlying data is unpredictable, and the fraud detection performance is susceptible to the amount of exploration.
 
\begin{figure}[t!]
\centerline{
      \includegraphics[width=\linewidth]{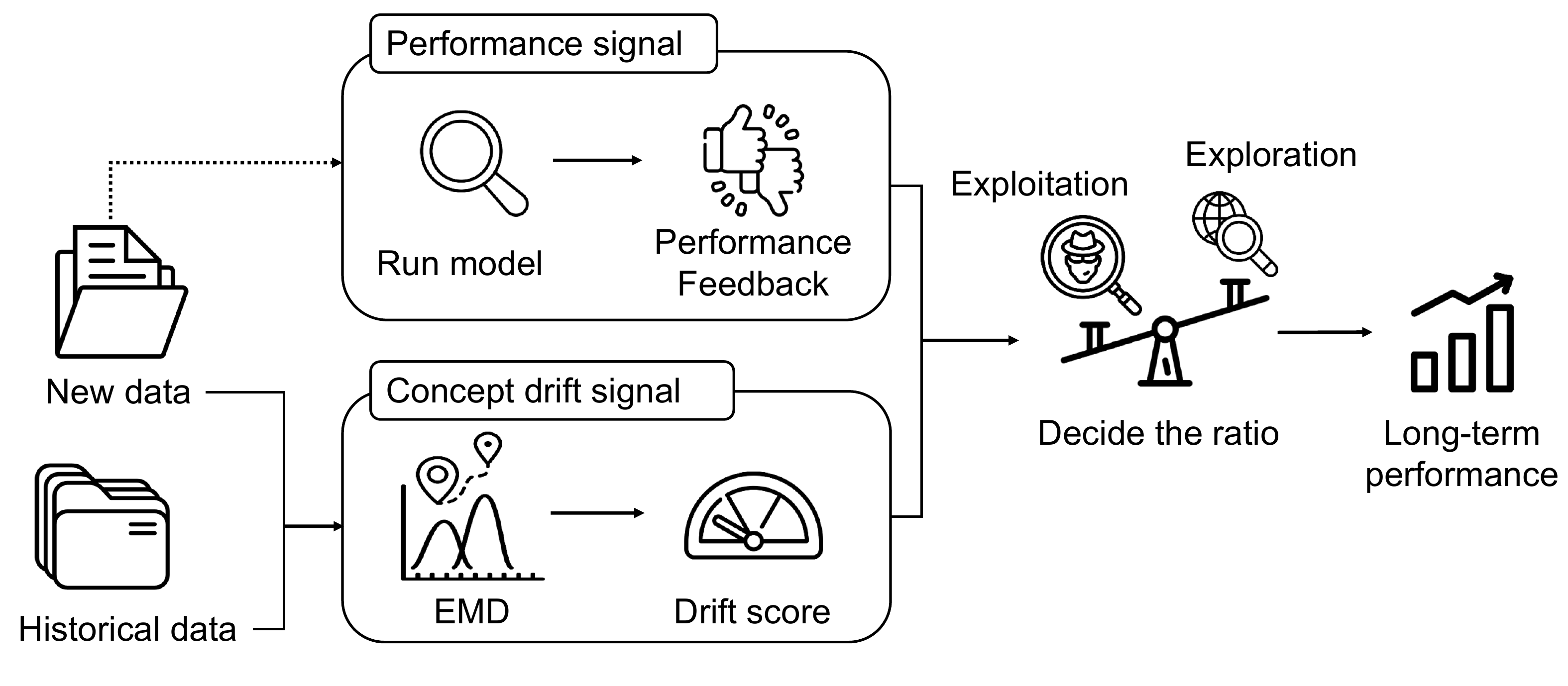}}
      \caption{\ours{} uses two signals for model sustainability.}
\label{fig:concept}
\end{figure}

This research introduces an Adaptive Drift-Aware and Performance Tuning (\ours{}) method to decide the balance between exploration and exploitation in detecting customs frauds. We measure the amount of concept drift to set a baseline of how much exploration should be used. In addition, we incorporate a performance signal to account for the fitness of the current model to the latest data. 
We utilize a multi-armed bandit framework with each arm corresponding to an exploration ratio. The changes in data distribution are measured and used to select a range of candidate arms for further consideration. Then, the model considers the historical performance of the candidate arms for its final decision. 
Fig.~\ref{fig:concept} illustrates the concept of the method.

To evaluate \ours{}, we benchmark its performance against the optimal strategy, which determines the exploration ratio by accessing the whole dataset in advance. This setting is unrealistic but provides a meaningful ``gold standard" for the model assessment. Experiments on multi-year and million-item import declarations from the four countries report that customs fraud detection system with \ours{} operates sustainably for a long period. Further analysis on the import declarations with previously unencountered characteristics demonstrates that \ours{} is effective in detecting new types of frauds. Finally, we demonstrate the importance of concept drift by examining its correlation with the model performance. 



For risk management in customs administrations, \ours{} provides two benefits. It can automatically balance the exploration and exploitation rate, without the need for external information, such as other models' performance or extensive hyperparameter tuning. This will save time and guarantee efficiency in the customs workflow. It can also act as a warning trigger for the offices and inform that trade pattern changes. 

\section{Related Work}
\subsection{Customs fraud detection} Previous works in the customs fraud detection domain suggest solutions in two major directions. The first body of work proposes approaches for exploring new cases, from the simple but intuitive random examination~\cite{han2014kcs} to more advanced active learning solutions using uncertainty~\cite{houlsby2011bald} and diversity~\cite{sener2018coreset, Ash2020badge}. The second group of work focuses on exploiting based on the previously obtained knowledge, which includes adapting heuristic approaches and off-the-shelf machine learning techniques, such as XGBoost~\cite{chen2016xgboost}, SVM~\cite{vanhoeyveld2020belgian}, or sophisticated deep-learning-based methods like DATE~\cite{kim2020date}. Inspecting some random items helps to improve the fraud detection performance even it sacrifice some known frauds~\cite{kim2020take}. Determining the exploration ratio in the human-in-the-loop fraud detection problem is under-explored, which is the goal of this paper. 

\subsection{Concept drift detection} Due to the proven effectiveness of concept drift for data analysis and performance of machine learning algorithms, there is extensive literature targeting this problem. Lu et al. classified these approaches~\cite{lu2019conceptdrift} in three categories; error-based drift detection~\cite{ddm}, data distribution-based drift detection~\cite{lu201411}, \cite{du2014informationentropy}, and multiple hypothesis test drift detection~\cite{alippi2008justintime}.
Error-based approaches refer to the performance of the classifier, while the distribution-based methods utilize the concept of dissimilarity between new and historical data distributions~\cite{wang2015concept}. 
The third category uses statistical tests to decide the final concept drift value. 
In this work, we conduct distribution-based drift detection analysis and further use this result to decide the exploration ratio of a fraud detection system.

\subsection{Multi-armed bandit problem}
The Multi-Armed Bandit Problem (MABP) addresses the dilemma of balancing between exploration and exploitation. The goal of the problem is to decide the best strategy for achieving a higher reward. Specifically, the decision is made between either relying on the previously discovered information or giving a chance for exploring new choices. 
A representative approach for dynamically adjusting the exploration-exploitation ratio is the exponential weighted framework. RP1 algorithm leverages an online learning mechanism using this framework to dynamically tune the ratio~\cite{rp1}. MABP has two settings, stochastic and adversarial. They differ by the reward model being applied \cite{slivkins_2019}. This work belongs to the adversarial bandits, where the agent makes no assumption regarding the reward generation process. A representative approach for the adversarial bandits in an online setting is the EXP3 algorithm \cite{exp3}. It works by assigning weights for each action, picking an action based on assigned weight, and then updating its weight based on the observed payoff. EXP3.S~\cite{exp3s} is a variant of the EXP3 algorithm, which adds regularization to ease arm switching. We improve upon their method by additionally accounting for the effect of concept drift.

\section{Problem Definition}
\label{sec:problemdefinition}

\subsection{Customs selection problem}

The human-in-the-loop customs selection problem, introduced in \cite{kim2020take}, is formally described as follows: 


At each timestamp $t$, customs receive a batch of items $\mathcal{B}_t$ from trade flows $\mathcal{B}$. Based on a strategy $f$ trained with labeled data $X_t$, customs officers select a batch of items $\mathcal{B}_t^S$ for manual inspection. After inspection, the newly annotated results are added into the training data for the next iteration $X_{t+1}$. 
The goal is to devise a strategy \textit{$f^*$} that maximizes the precision and revenue in the long term:

\begin{equation}
f^* = \operatorname*{argmax}_f{\sum_{t}{m(\mathcal{B}_t^s(f))}},
\end{equation}
where $m$ is the key performance index for the fraud detection system such as precision and revenue. 

\subsection{Customs selection strategies}
\label{sec:def:strategy}
A customs selection strategy $f$ generally falls into one of the two categories:

\begin{itemize}
    \item Exploitation strategy $f_{explore}$: Selecting likely fraudulent and lucrative items. It guarantees customs to collect tariffs, ensuring revenue.
    \item Exploration strategy $f_{exploit}$: Selecting unseen items. Adequately exploring diverse trade items allows to detect novel frauds and prevent confirmation bias.
\end{itemize}

A mixture of exploitation and exploration strategies---a hybrid approach is proven to effectively maintain long-term performance~\cite{kim2020take}. The hybrid approach selects $k|\mathcal{B}_t^S|$ from $f_{explore}$ and $(1-k)|\mathcal{B}_t^S|$ items from $f_{exploit}$, with $k$ being the constant exploration ratio. The concept of hybrid strategy using a constant amount of exploration is similar to the $\epsilon$-greedy algorithm~\cite{slivkins_2019}. However, unlike $\epsilon$-greedy, the hybrid approach can support any exploration strategy $f_{explore}$ other than random exploration.

\subsection{Finding the best exploration ratio}
In this paper, we extend the problem to the case where $k$ is no longer a constant. The algorithm aims to find the best exploration ratio $k_t$ for every timestamp $t$. The problem can be formally defined as follows: Given the exploration strategy $f_{explore}$ and exploitation strategy $f_{exploit}$, choose the exploration ratio $k_t$ for current timestamp $t$.

\begin{equation}
k_t^* = \operatorname*{argmax}_{k_t}{\sum_{t}{m(\mathcal{B}_t^s(k_t;f_{explore},f_{exploit}))}}.
\end{equation}

\section{Method}
The performance of the machine learning model for customs fraud detection is sensitive to the degree of exploration \cite{kim2020take}. However, there is little work regarding how to manage this trade-off effectively. To make a robust selection model, we consider an exploration-exploitation trade-off in our proposed method, updating the ratio according to two signals; model performance and concept drift.

\begin{figure*}[t!]
\centerline{
      \includegraphics[width=0.9\linewidth]{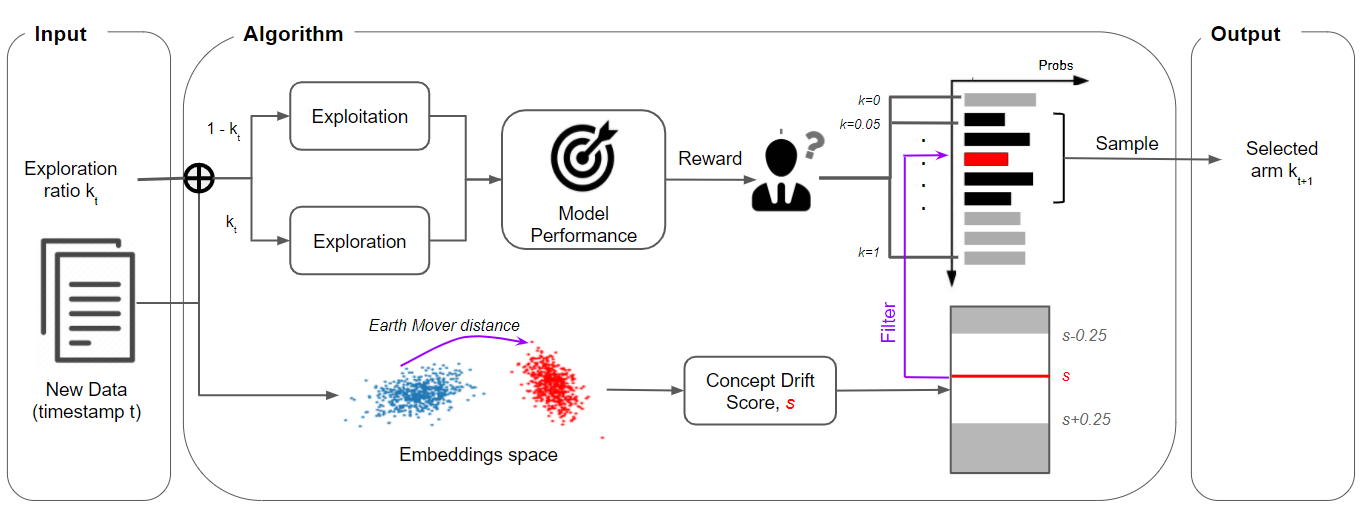}}
      \caption{Detailed workflow of the model. The performance signal helps the multi-armed bandit algorithm to update the probabilities of choosing each arm. Meanwhile, the drift signal suggests the exploration ratio be in a certain range. The arms lying outside the range are eliminated, and the probabilities are normalized for final arm selection.}
\label{fig:model}
\end{figure*}

Those two signals relate to two concepts: the multi-armed bandit problem, where actions are chosen based on rewards, and concept drift when data shift its distribution over time. When the performance drops or the underlying data distribution changes, the model adjusts the exploration level, 
From MABP's perspective, the probability of choosing each arm is updated by the historical result. Meanwhile, concept drift narrows down the choice to a smaller set of arms, acting as a filter. Figure~\ref{fig:model} illustrates this concept.

\subsection{Performance signal with multi-armed bandit}

The multi-armed bandit problem is a classic example of solving the exploration-exploitation dilemma. The performance of the model can be used as the reward to update the bandit and guiding the bandit to the region of higher performance. Due to the arbitrary nature of the incoming trade flow, we cannot make any well-defined statistical assumptions about the generation of rewards, which corresponds to the adversarial setting~\cite{exp3}. 

EXP3 (stands for exploration, exploitation, exponentiation) is a widely used algorithm for adversarial bandit~\cite{alessiardo2015exp3}. This framework assumes a set of actions (arms) $A= \{a_1,a_2,…,a_n\}$ which essentially corresponds to different exploration ratios $\{k_1,k_2,…,k_n\}$. At the beginning of each timestamp, the learner must choose an action $a_t$ to minimize the cumulative regret. The regret of each timestamp is defined as the reward difference between the optimal action and the actual action taken.

\begin{equation}
Regret = \sum_{t=0}^T {\max_{a \in A}{R(a, B_t) - R(a_t,B_t)}}
\end{equation}

The framework maintains a guiding distribution $p_t$ over the set of actions $A$ and uses it to sample the action. The algorithm will receive a reward $R$ (preferably for each action), and the guiding distribution is updated by decreasing the weights of ‘bad’ actions and exponentially raising the weights of ‘good’ actions. Since we only get the reward for the taken action, we use an unbiased estimator \cite{bubeck2012stochasticadversartial}: 

\begin{equation}
\hat{R}_t(a_i) = \dfrac{R_t(a_t) \mathds{1}_{(a_i=a_t )}}{p_{t}(a_t)}.
\end{equation}

The distribution is updated by
\begin{equation}
p_{t+1}(a_i) = p_{t}(a_i) e^{\eta \hat{R_t}(a_i)},
\end{equation}
where $\eta$ is the learning rate.
The reward $R$ is usually normalized to be bounded by 1. EXP3 is built to ﬁnd the best arm on the entire run and it attains an asymptotic regret $O(\sqrt{n})$, where $n$ is the number of rounds.

A variant of EXP3, namely EXP3.S \cite{alessiardo2015exp3}, adds a regularization term to the update function to ease arm switches and better adapt to a non-stationary environment. It achieves a similar asymptotic regret bound.  

To adapt to our setting, each arm will correspond to an exploration ratio. We use a 21-arm framework corresponding to 21 ratios ranging from 0 to 1 with a step size of 0.05. In order to maintain a robust performance, precision is given as the feedback, such that the exponential-weighted framework could timely respond in the event of a performance drop. The reward in our model is the current precision compared to the weighted averaged precision of all rounds. A discount factor of $\gamma$ is used to put more weight on recent results. We call this strategy Adaptive Performance Tuning (APT), which is formally presented in Algorithm~\ref{alg:mab}.
\begin{algorithm}[bth]
\SetNoFillComment
\begin{flushleft}
 \textbf{Parameters:} Learning rate $\eta$, randomness $\epsilon \in [0, 1]$ and regularization factor $\alpha > 0$, discount factor $\gamma$.\\

 \textbf{Initialization:} $w_0 = [1, 1,\ldots, 1]$ \\
 
 \SetAlgoLined
 \For{$t = 1, 2, 3,\ldots, T$}{
    1. Set selection probability:
        $$p_t(a_i) = \dfrac{\epsilon}{k} + (1-\epsilon)\dfrac{w_t(a_i)}{\sum_i w_t(a_i)}$$
    
    2. Draw an arm $a_t$ according to the probabilities $p_t$, and get the corresponding precision $\pi_t$

    3. Calculate weighted average performances of all round, with discount factor $\gamma$:

    $$\overline{\pi} = \dfrac{\pi_t + \gamma \pi_{t-1} +\ldots+ \gamma^{t-1}\pi_1}{1 + \gamma +\ldots+ \gamma^{t-1}}$$
    
    4. Calculate the reward :
    
    $$R_t = \dfrac{\pi_t - \overline{\pi}}{\pi_t}$$
    
    5. Update:
        $$\hat{R}_t(a_i) = \dfrac{R_t(a_t) \mathds{1}_{(a_i=a_t )}}{p_{t}(a_t)}$$
        $$w_{t+1}(a_i) = w_{t}(a_i) e^{-\eta \hat{R_t}(a_i)} + \dfrac{e\alpha}{k} \sum_i w_t(i)$$
 }
 \caption{Adaptive Performance Tuning algorithm}
 \label{alg:mab}
\end{flushleft}
\end{algorithm}

\subsection{Measuring concept drift by optimal transport}
The second way to decide the exploration rate is by measuring the concept drift. Optimal transport theory is used to measure the distribution difference between data snapshots. Optimal transport aims to find the most efficient way to move mass between distributions~\cite{villani2016ot}. The cost of moving a unit of mass between two positions is called the ground cost, and the objective is to minimize the overall cost of moving one mass distribution to another one. Wasserstein has formulated this problem as the following: \cite{villani2016ot}
\begin{equation}
W_p(\mu, \nu) = \left[\inf_{\gamma  \in  \Gamma (x, y)} \int \mathcal{D}(x, y)^p d\gamma(x, y) \right]^{\frac{1}{p}}
\end{equation}
with $\mu$ and $\nu$ are the marginal probability distributions of $X$ and $Y$ with realized values of $x$ and $y$. $\Gamma(x, y)$ denotes the set of all possible joint distributions between $X$ and $Y$. $\mathcal{D}$ is a distance function. $p$ is the power of the distance, which can be an integer.
 
Earth mover's distance (EMD) is a measure of the distance between two data distributions. It measures the minimum cost of turning one pile of distribution into the other. To bound EMD to have a specific range of measuring concept drift, we consider how much the optimal EMD value has improved from its upper bound, obtained by applying the norm inequality~\cite{villani2016ot}: 
\begin{align}
  W_1(\mu, \nu) &\leq \inf_{\gamma  \in  \Gamma (x, y)} \int [\mathcal{D}(x) + \mathcal{D}(y)] d\gamma(x, y) \nonumber
 \\ &\leq \int \mathcal{D}(x)d\mu (x) + \int \mathcal{D}(y)d\nu (y).
\end{align}
By dividing the left-hand side by the right-hand side, the concept drift score $s$ is obtained with the range of 0 to 1, allowing us to obtain an interpretable threshold. Especially in an online learning setting, the bounding is necessary since the distribution and values frequently change. The closer the value is to 1, the more statistically different the two data snapshots are. \looseness=-1

In the customs setting, we compare the distance between recent historical data and incoming data to measure the concept drift. In our online setting, we use data from the recent four weeks as the validation data. Hence, it becomes handy to use the validation set as the historical data. We encode each import declaration to 16 dimensions by using a transaction encoder~\cite{kim2020date}, and the concept drift is measured by calculating the normalized earth mover's distance between the two embedding sets \cite{flamary2021pot}. Equal probability is assigned to each data point, and the bootstrap technique is used to reduce training time and obtain robust results. The power of the earth mover's distance is chosen as 1, with Euclidean distance as the distance function~\cite{bonneel2011pot}. The result of normalized earth mover's distance can be thought of as the irreducible cost to turn one distribution to another or the new dataset's novelty from the original dataset and will be used as the ratio for data exploration.

We can use this concept drift score directly as the exploration ratio since if we have more concept drift, there are more patterns to discover and potentially need more exploration. We call this method Adaptive Drift-Aware (ADA) strategy.

\subsection{Adaptive Drift-Aware Performance Tuning algorithm}

APT makes its decision from a probability distribution spread over different exploration ratios, while ADA measures the concept drift and puts its bet on a single point in the unit interval. The proposed model, \ours{} (Adaptive Drift-Aware Performance Tuning), makes the most out of two strategies by choosing arms within a more targeted range. More specifically, after getting the concept drift score $s$, we only consider the arms corresponding to the exploration ratio inside the interval $[max(0, s - l), min(1, s + l)]$.
Only these arms are granted probabilities of being chosen by the APT algorithm, while all other arms outside of the region are given a probability of 0. The probabilities for all arms are then normalized and ready for the next iteration. We can consider this as applying a filter center at $s$, with $2l$ as the window width. We set $l$ as $0.25$ to consider a reasonably wide range of arm candidates.

\section{Experiment}

For evaluation, we first review our experiments to answer the following questions:

\begin{itemize}
\item Identify the need for exploration (Q1): How much exploration do we need for maintaining a sustainable system? 

\item Effectiveness of the adaptive strategies (Q2): How well does \ours{} find the ideal exploration ratio?

\item Concept drifts analysis (Q3): Does \ours{} capture the concept drift effectively, and does concept drift substantially affect model performance?

\item Novel fraud detection (Q4): Is \ours{} effective in detecting frauds among novel trades?

\item Ablation analysis (Q5): How important is each component?

\end{itemize}

\subsection{Settings}
\subsubsection{Datasets}
\label{sec:experiments:settings:datasets}
For experiments, we employed item-level import declarations from four countries in Africa. The datasets span multiple years allowing us to observe concept drift. The import declarations include the item's free on board~(FOB) price, gross weight, quantity, tariff code, importing country, and handler information such as importer, declarant, customs office, and its estimated taxes. The dataset also contains inspected results of whether a transaction is a fraud or not. Due to the data confidentiality policy, we call these countries M, C, N, and T. Table.~\ref{tab:datastats} shows the statistics of the datasets. 

Customs administrations in the four countries conducted nearly 100\% of manual inspections of their imported goods. Since the data obtained so far was under a complete inspection, illicitness of the transaction and charged tariffs are accurately labeled at the single-goods level. But this practice is not sustainable, and the customs offices of these countries plan to reduce the inspection rate in the future. 

\begin{table}[h!]
\centering
\caption{Statistics of the datasets}
\label{tab:datastats}
\resizebox{\linewidth}{!}{%
\begin{tabular}{ l | r | r | r | r } \toprule
    Datasets &  Country \textsf{M} & Country \textsf{C} & Country \textsf{N} & Country \textsf{T}  \\ \midrule
    Periods & 2013--2016 & 2016--2019 & 2013--2017 & 2015--2019 \\ 
    \# imports & 0.42M & 1.90M & 1.93M & 4.17M  \\ 
    \# importers & 41K & 9K & 165K & 133K  \\ 
    \# tariff codes & 1.9K & 5.5K & 6.0K & 13.4K  \\ 
    GDP per capita & \$400 & \$1,500 & \$2,200 & \$3,300 \\
    Illicit rate &  1.6\% & 1.7\%   & 4.1\% & 8.2\%  \\ 
    \bottomrule
\end{tabular}
}
\end{table}

\subsubsection{Long-term simulation setting}
\label{sec:experiments:settings:scenario}

We conduct experiments to check the ability of our proposed \ours{} method to choose the exploration ratio that maintains the customs selection model in the long run. We used the simulation setting proposed in \cite{kim2020take}, where a selection model is deployed, updated, and maintained for multiple years. A limited amount of annotated training data is given to initialize the model. For each timestamp, the method updates the ratio between exploration-exploitation. With the ratio, the model selects a batch of items from the incoming import declarations stream. The custom officers manually inspect these items and their fraud labels are obtained. The model is re-trained with the accumulated training set, and the most  recent four weeks of data are used to validate the model. The performance for each timestamp is recorded to evaluate the method. The average performances over the whole period, last two years, last one year, and last six months are additionally reported to summarize the performance.


\subsubsection{Evaluation metrics}
\label{sec:experiments:settings:metrics}
If $n\%$ of all import declarations are inspected, the performance of the customs selection model in the online setting is measured by two metrics introduced in \cite{kim2020take}: Norm-Precision@n\% and Norm-Revenue@n\%. They are the precision and revenue evaluated on the selected items, divided by the maximum achievable value. Precision indicates the proportion of fraud cases among the inspected items. Revenue indicates the proportion of revenue secured by examining the set of items, which puts more weight on the lucrative items. The normalization mitigates the change in difficulty of the dataset at different times. 

\begin{figure*}[t!]
    \begin{subfigure}[b]{0.42\linewidth}
        \centering\captionsetup{width=.95\linewidth}%
\includegraphics[trim={0 0 5.5cm 0},clip,width=\linewidth]{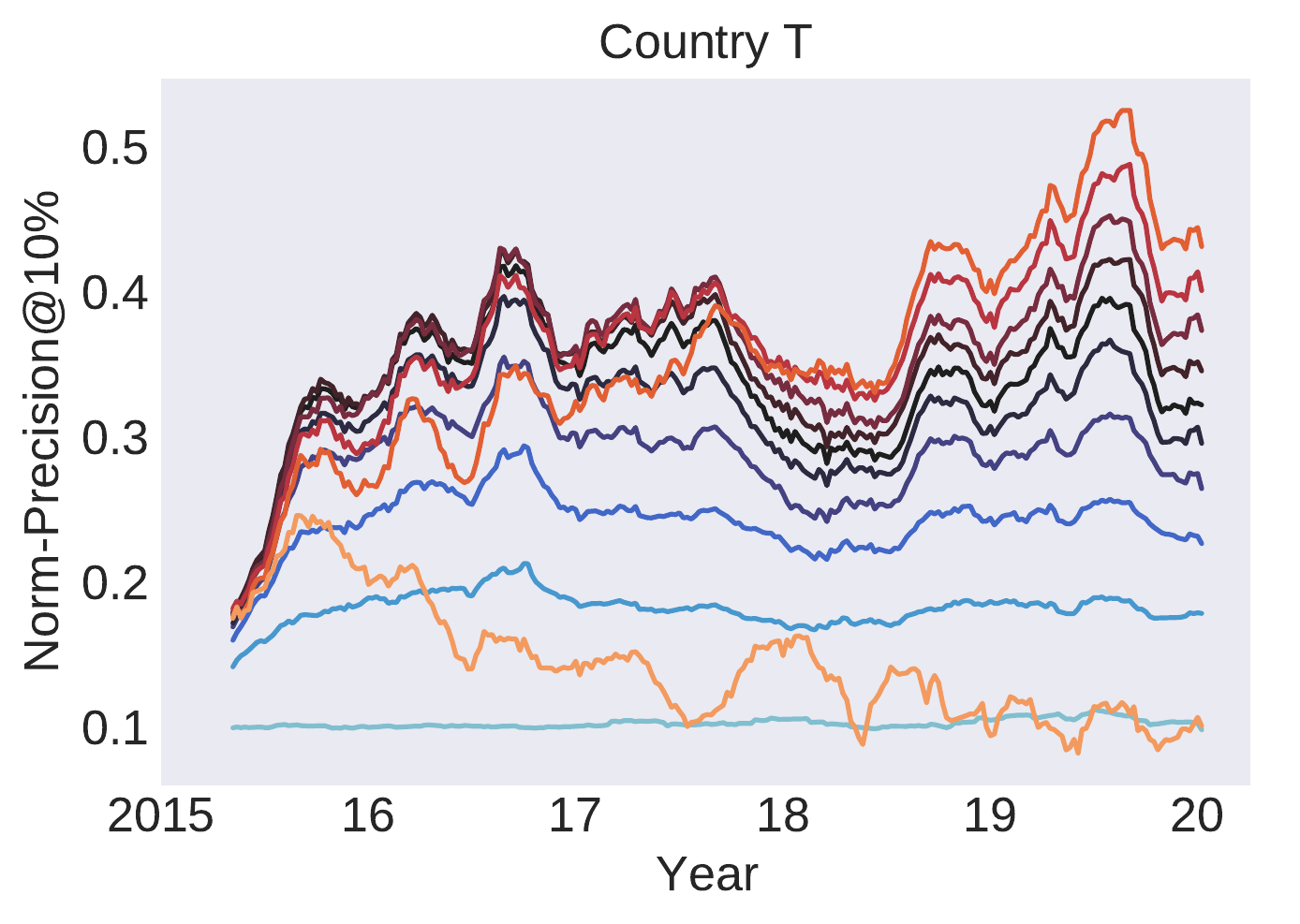}
        \caption{In country \textsf{T}, the performance of the full exploitation strategy drops over time but injecting exploration keeps the performance robust. 10\% of exploration yields the best performance.}
    \end{subfigure}
    \begin{subfigure}[b]{0.42\linewidth}
        \centering\captionsetup{width=.95\linewidth}%
        \includegraphics[trim={0 -0.2cm 5.5cm 0},clip,width=\linewidth]{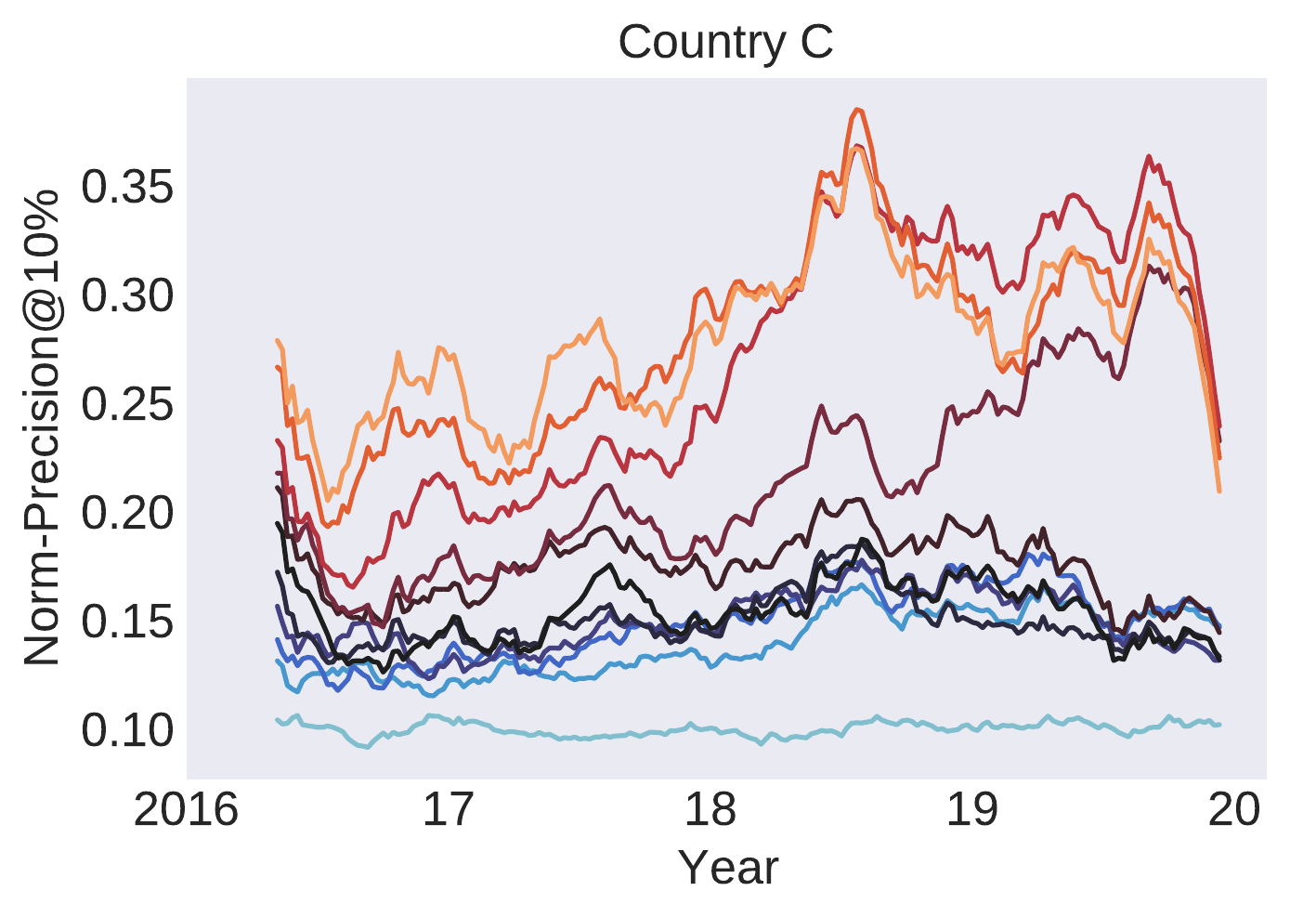}
        \caption{In country \textsf{C}, 20\% of exploration yields the best performance. This setting underperforms full exploitation at the beginning, but the performance improved as time goes.}
    \end{subfigure}
     \begin{subfigure}[b]{0.15\linewidth}
        \centering
        \includegraphics[trim={15cm -3cm 0 0},clip,width=\linewidth]{figures/hybrid-real-c-norm-precision.pdf}
    \end{subfigure}
    
    \begin{subfigure}[b]{.42\linewidth}
        \centering\captionsetup{width=.95\linewidth}%
\includegraphics[trim={0 0 5.5cm 0},clip,width=\linewidth]{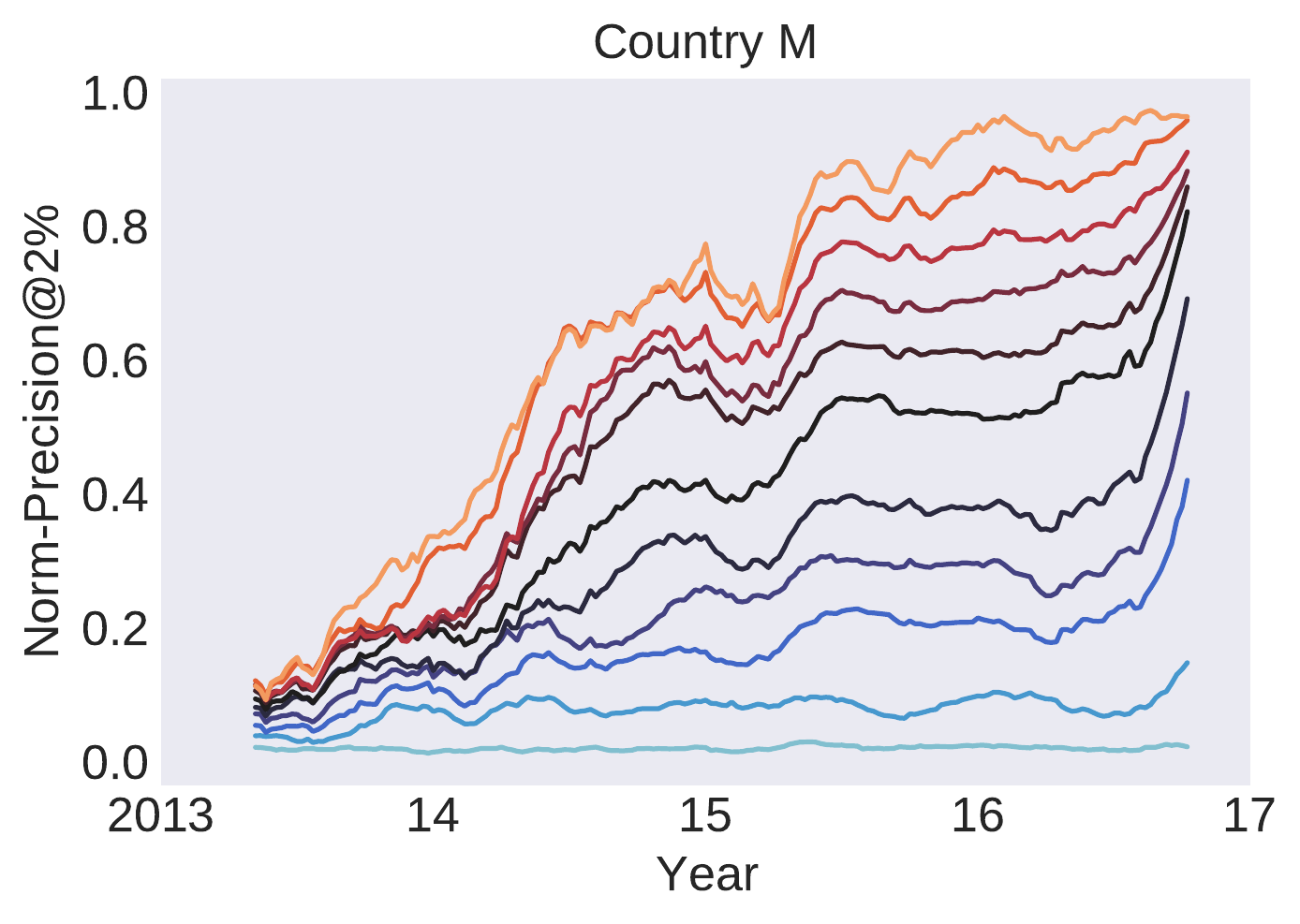}
        \caption{In country \textsf{M}, fully exploitation yields the best performance.}
    \end{subfigure}
    \begin{subfigure}[b]{.42\linewidth}
        \centering\captionsetup{width=.95\linewidth}%
        \includegraphics[trim={0 0 5.5cm 0},clip,width=\linewidth]{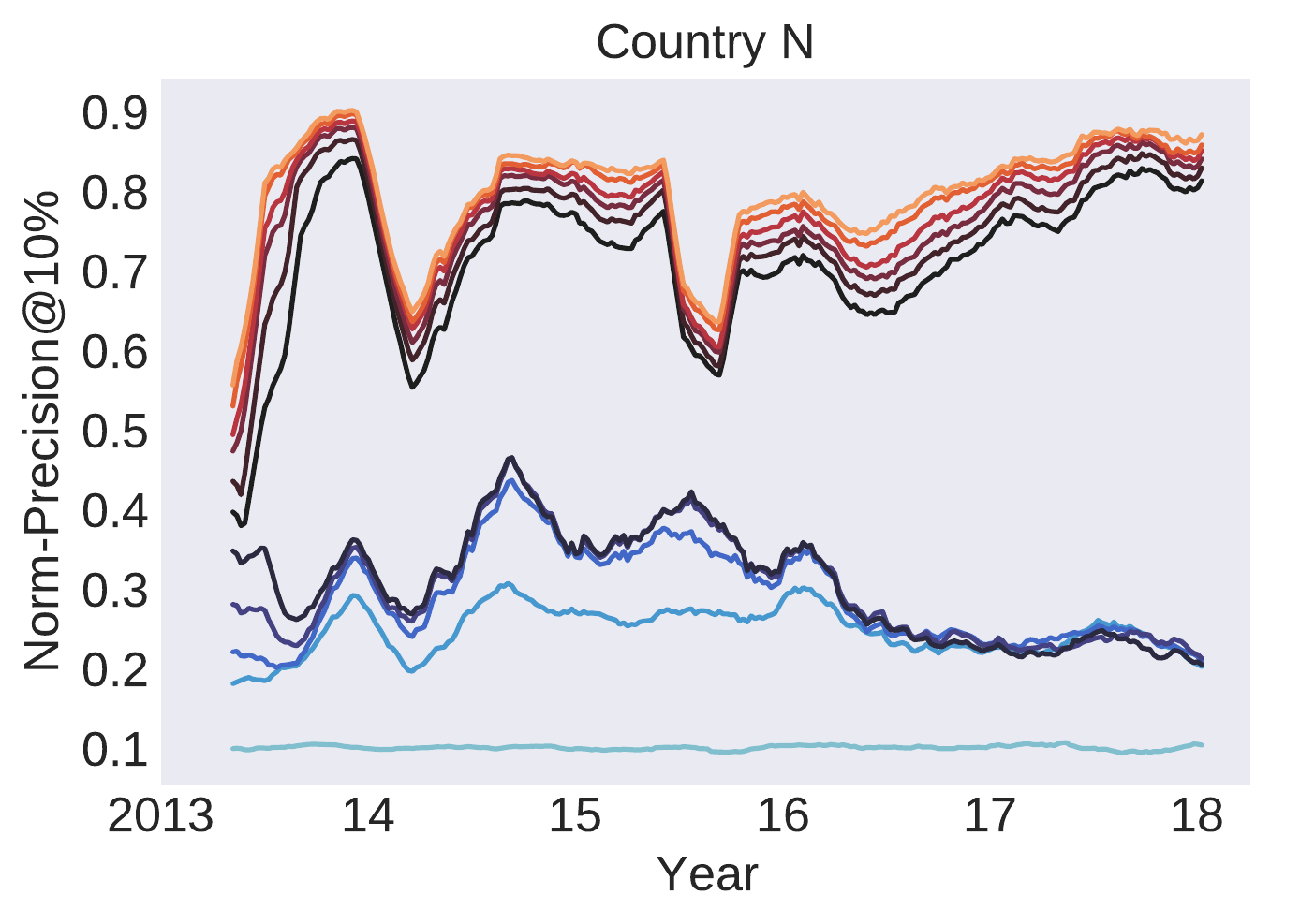}
        \caption{In country \textsf{N}, fully exploitation yields the best performance.}
    \end{subfigure}
    \begin{subfigure}[b]{0.15\linewidth}
        \centering
        \includegraphics[trim={15cm -2cm 0 0},clip,width=\linewidth]{figures/hybrid-real-c-norm-precision.pdf}
    \end{subfigure}
    
    \caption{\textbf{Performance of hybrid with different exploration ratios in 4 countries (T, C, M, N)}. The ratio between exploration and exploitation greatly affects the overall performance in all countries. Adding some exploration helps  to improve the performance in data T and C, while fully exploitation strategy works best in data M and N. Results suggest that each country has its own suitable exploration rate.}
    \label{fig:hybrid}
\end{figure*}

\subsubsection{Training Details}
We use XGBoost (GBDT)~\cite{chen2016xgboost} as our exploitation strategy and random \cite{han2014kcs} as our exploration strategy. For Exp3.S, we set the learning rate as 3.0, randomness as 0.1, regularization weight as 0.001, and discount factor as 0.9. The inspection rate is 10\%, except for country M we use 2\%. This is because data M has a low fraud rate and the fraud pattern is relatively easy so that many algorithms reach 100\% precision if an inspection rate is kept at 10\%. Each experiment is run five times and the averaged results are reported. To smooth out short-term fluctuations and highlight longer-term trends, we show the moving average of 14-weeks in the following figures.

\subsection{Performance Evaluation}
\subsubsection{Identify the need for exploration (Q1)}

We first identify the need for exploration by simulating the customs selection process in four countries. Assuming that each country faces a different amount of concept drift, we first run hybrid strategies (Sec.~\ref{sec:def:strategy}) with different amounts of exploration to determine the most appropriate exploration ratio for each country. We record the precision and revenue over time by varying the exploration ratio ranging from 0 to 1. The precision trend is reported in Fig.~\ref{fig:hybrid}.
Results reveal that each country has its own suitable exploration rate. For country C and country T, 20\% and 10\% of exploration yield the best performance, respectively. Country M and N do not experience significant concept drift, and the full exploitation strategy works the best. Therefore, our main experiment will focus on the dataset with the concept drift, namely country C and T.

In country T, the performance of the full exploitation strategy degrades over time. However, injecting some explorations into the strategy keeps the performance more robust as time passes. Model performance is sensitive to the exploration rate. Models with 10--30\% exploration achieve the highest precision, while models with 50--70\% exploration perform moderately. Fully exploration or fully exploitation strategies performance is worse than any mixtures, with a difference of more than 0.3 at some timestamps.

For country C, full exploitation performed the best at first. However, since 2018, the 20\% exploration rate takes the lead in performance, followed by the 10\% exploration rate. The strategies with 60\% or more exploration perform far worse than others. The best ratio does not stay the same over time but also depends on the period in consideration. The exploration rate of 20\% is ultimately the best strategy.  

For brevity, we refer to the best performing hybrid in the last six-month period as the \textit{oracle}. Performance statistics of oracle are reported in Table.~\ref{tbl:main}.

\subsubsection{Effectiveness of the adaptive strategies (Q2)} 
\begin{figure*}[t!]
    \begin{subfigure}[b]{0.41\linewidth}
        \centering\captionsetup{width=.95\linewidth}%
\includegraphics[trim={0 0 5.0cm 0},clip,width=\linewidth]{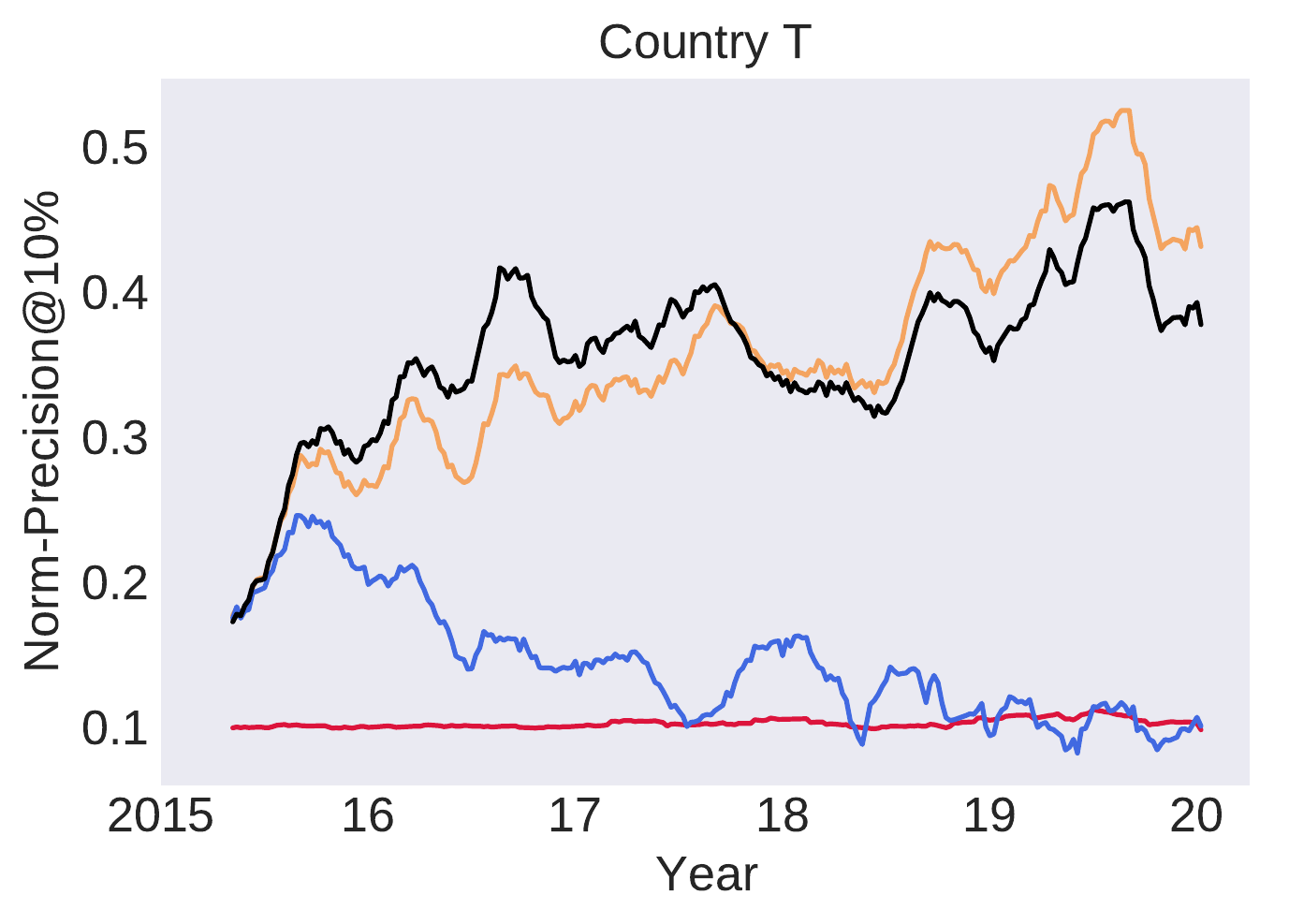}
        \caption{In country \textsf{T}, \ours{} performance is approaching the performance of with the oracle.}
    \end{subfigure}
    \begin{subfigure}[b]{0.42\linewidth}
        \centering\captionsetup{width=.95\linewidth}%
        \includegraphics[trim={0 0cm 5.0cm 0},clip,width=\linewidth]{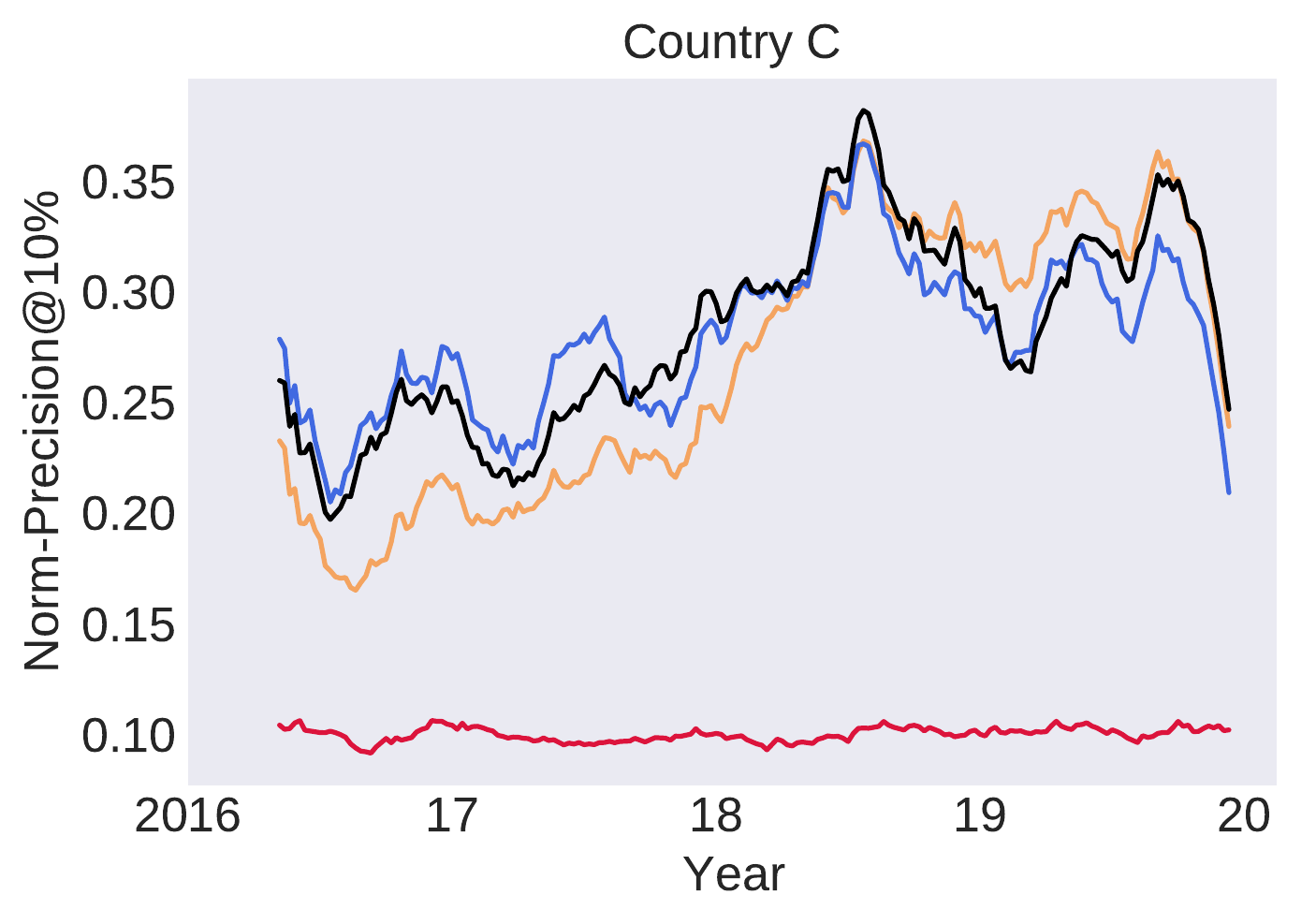}
        \caption{In country \textsf{C}, \ours{} performs similarly with the oracle.}

    \end{subfigure}
     \begin{subfigure}[b]{0.15\linewidth}
        \centering
        \includegraphics[trim={14.5cm -1.4cm 0 0},clip,width=\linewidth]{figures/rada-real-c-norm-precision.pdf}
    \end{subfigure}
    
    \begin{subfigure}[b]{.42\linewidth}
        \centering\captionsetup{width=.95\linewidth}%
\includegraphics[trim={0 0 5.5cm 0},clip,width=\linewidth]{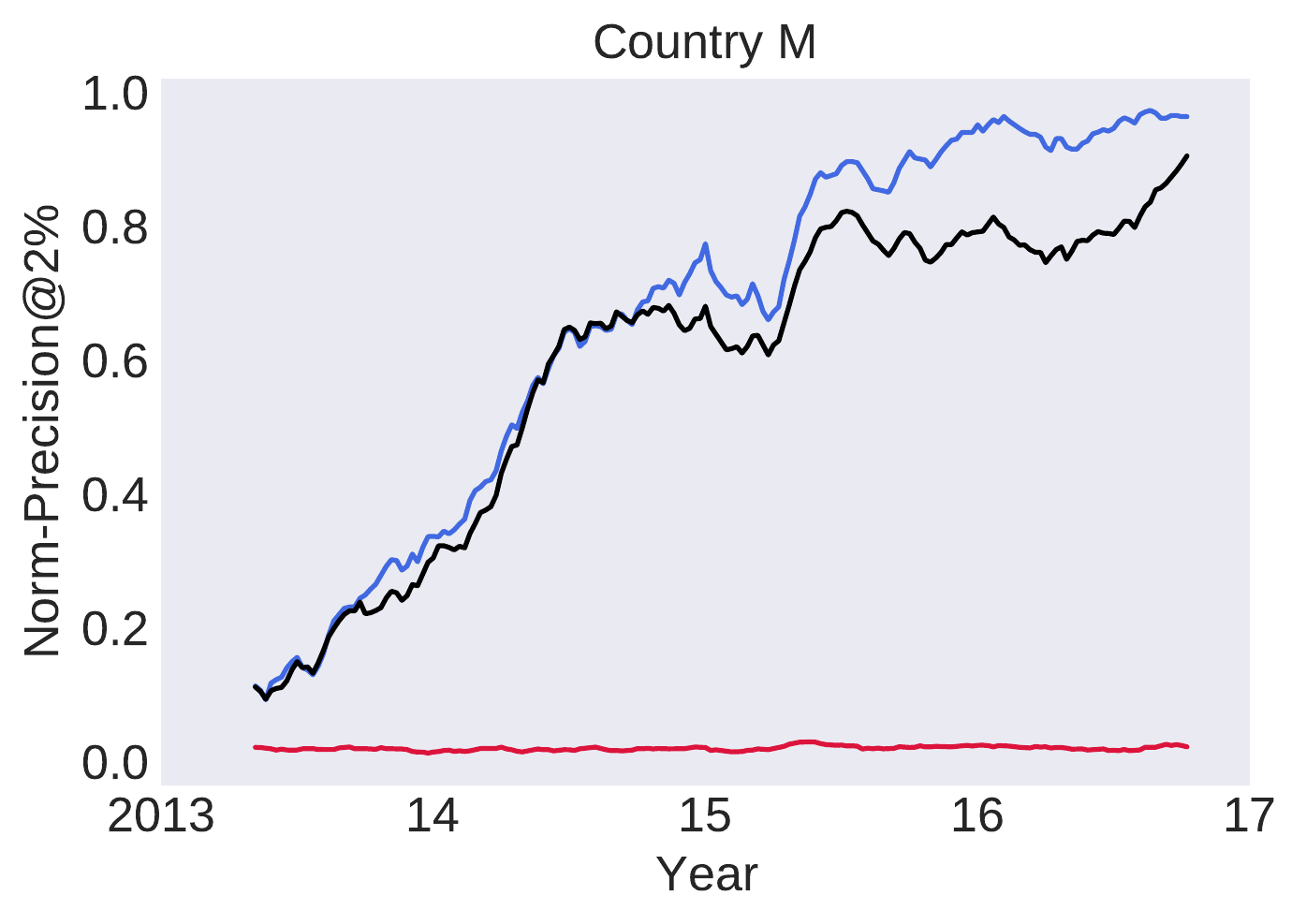}
        \caption{In country \textsf{M}, \ours{} performance is approaching the performance of with the oracle.} 
    \end{subfigure}
    \begin{subfigure}[b]{.42\linewidth}
        \centering\captionsetup{width=.95\linewidth}%
        \includegraphics[trim={0 0 5.5cm 0},clip,width=\linewidth]{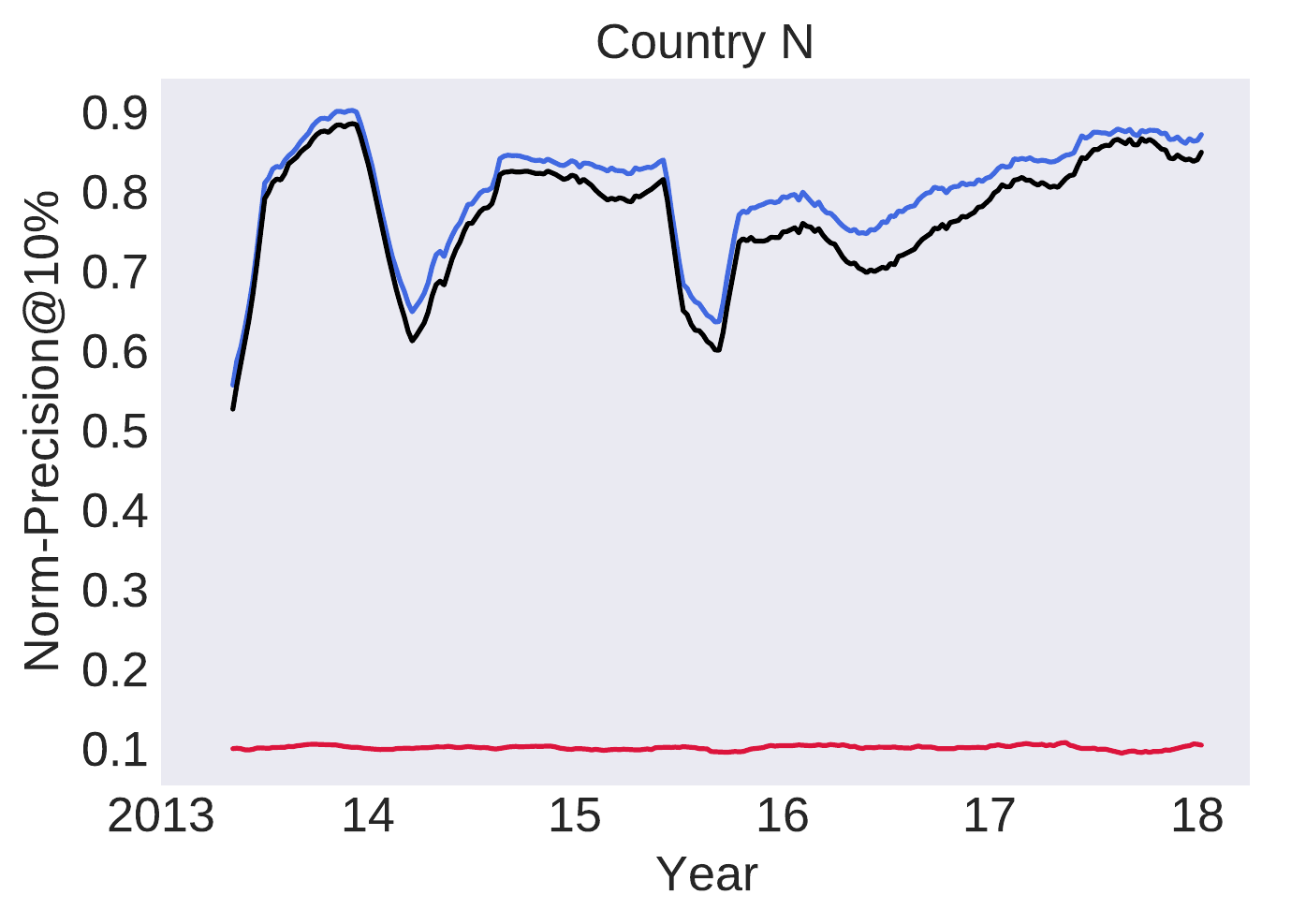}
        \caption{In country \textsf{N}, \ours{} performance is approaching the performance of with the oracle.}
    \end{subfigure}
    \begin{subfigure}[b]{0.15\linewidth}
        \centering
        \includegraphics[trim={14.3cm -1.85cm 0.42cm 0},clip,width=\linewidth]{figures/rada-real-n-norm-precision.pdf}
    \end{subfigure}
    
    \caption{\textbf{Performance of \ours{} in 4 countries (T, C, M, N)} compared to the oracle, fully exploration and fully exploitation. In all countries, \ours{} performance approaches that of the oracle.}
    \label{fig:rada}
\end{figure*}

Through this experiment, we confirmed how much the risk management system improved when the adaptive exploration rate is determined by the proposed \ours{} method. Precision and revenue are reported in Table.~\ref{tbl:main}. Precision trends compared with some baselines are shown in Fig.~\ref{fig:rada}.

In country C, \ours{} follows the similar trend with the fully exploitation model, which was the best performing one until the middle of 2018. From that point, hybrid with 20\% exploration (oracle) outperforms all hybrid baselines and fully exploitation strategy. Likewise, \ours{} successfully adapt to the concept drift and eventually follows the same trend and performance of the oracle. According to Table.~\ref{tbl:main}, \ours{} was the most effective when all timestamps are considered.
In country T, \ours{} has outperformed the oracle in 2016--2017 period. In the next period, \ours{} performed closely to the oracle with similar trends. On average, \ours{} has similar average precision and revenue with the best hybrid strategy, with more stable performance through more than five years. \ours{} outperforms 9 out of 11 hybrid models. In country M and N, \ours{}'s performance approaches that of the best hybrid. Out of 11 hybrids, \ours{} outperform 7 models in country M and 8 models in country N.


From this experiment, we find that \ours{} makes the informed decision for every timestamp so that it achieves comparable performance to the best hybrid method at each point. Compared to the best performing hybrid, oracle, \ours{} even showed higher performance in some time periods.
With the accumulated data, it was possible to experiment with multiple hybrids with various setups, but it is infeasible to do this in real-world scenarios. In comparison, \ours{} can adaptively update the ratio based on new data immediately and does not require any hyperparameter tuning, showing a better practice for online setting.

\subsubsection{Concept drift analysis (Q3)}

\begin{figure*}[t!]
   \begin{minipage}{0.3\textwidth}
     \centering
     \includegraphics[width=0.90\linewidth]{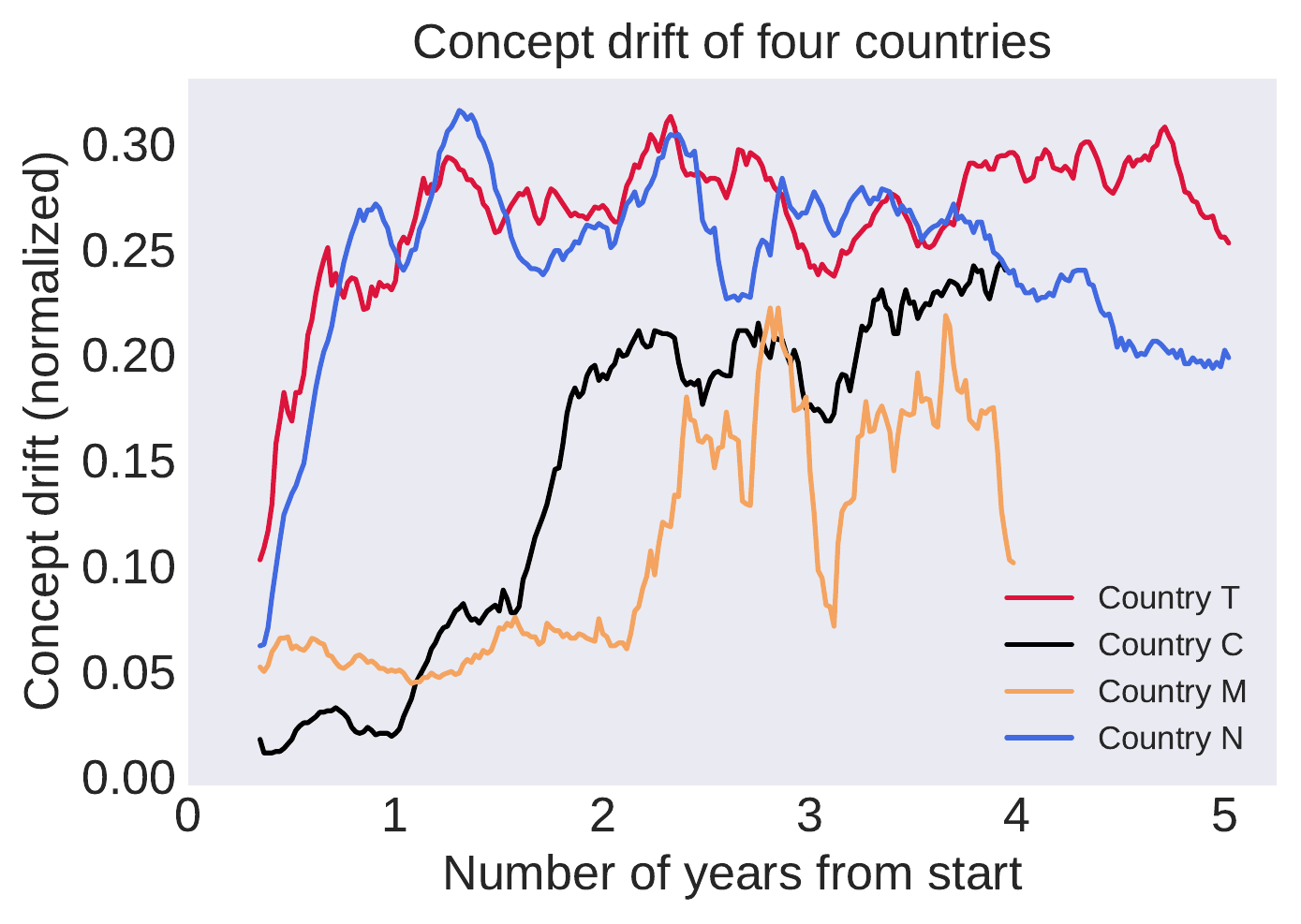}
     \caption{Concept drift score measured by optimal transport on the data from four countries.}
     \label{fig:cd-all}
   \end{minipage}
   \hfill
   \begin{minipage}{0.68\textwidth}
     \centering
    \begin{subfigure}[b]{0.48\linewidth}
        \centering\captionsetup{width=.95\linewidth}%
        \includegraphics[width=\linewidth]{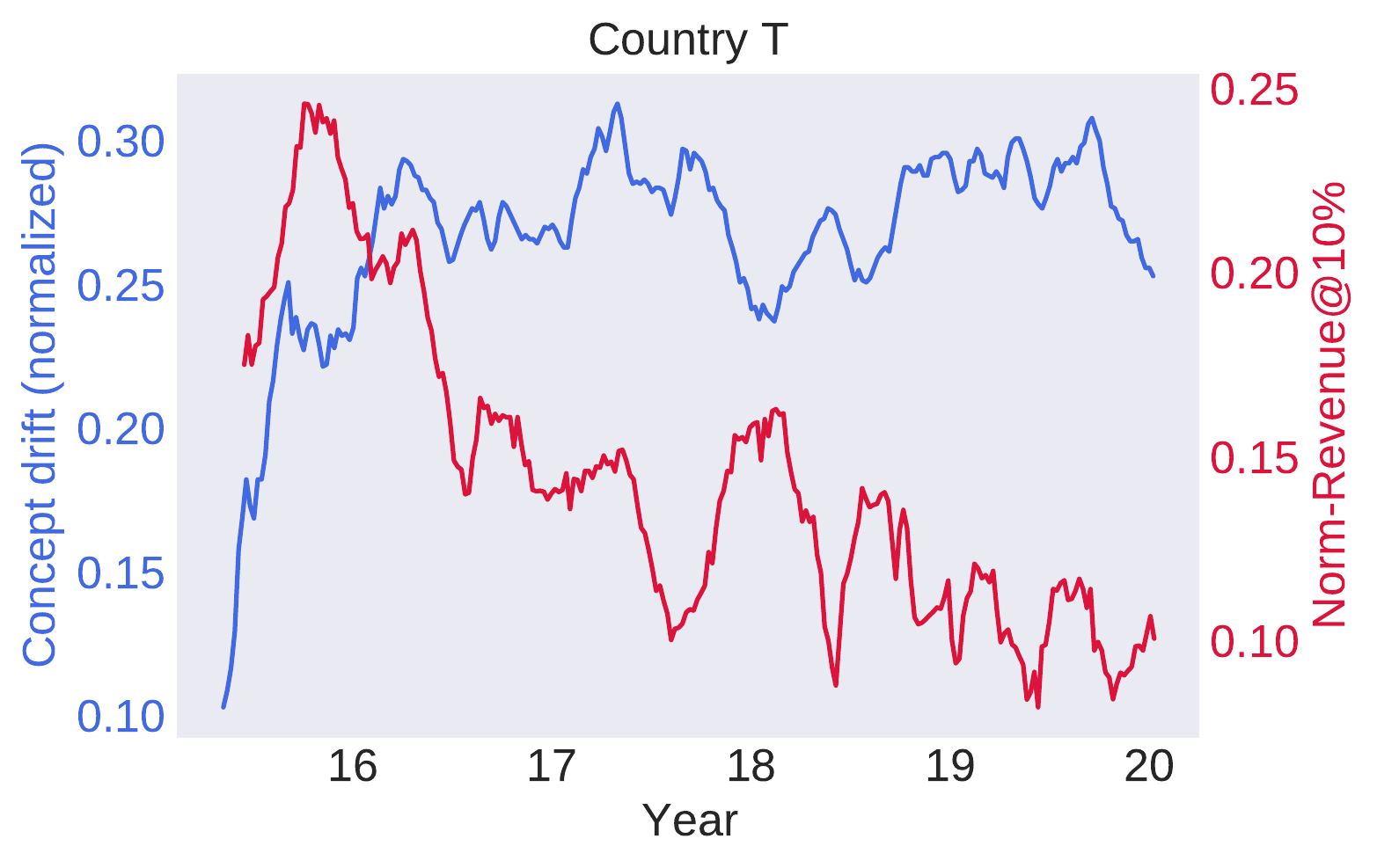}
    \end{subfigure}
    \begin{subfigure}[b]{0.49\linewidth}
        \centering\captionsetup{width=.95\linewidth}%
        \includegraphics[width=\linewidth]{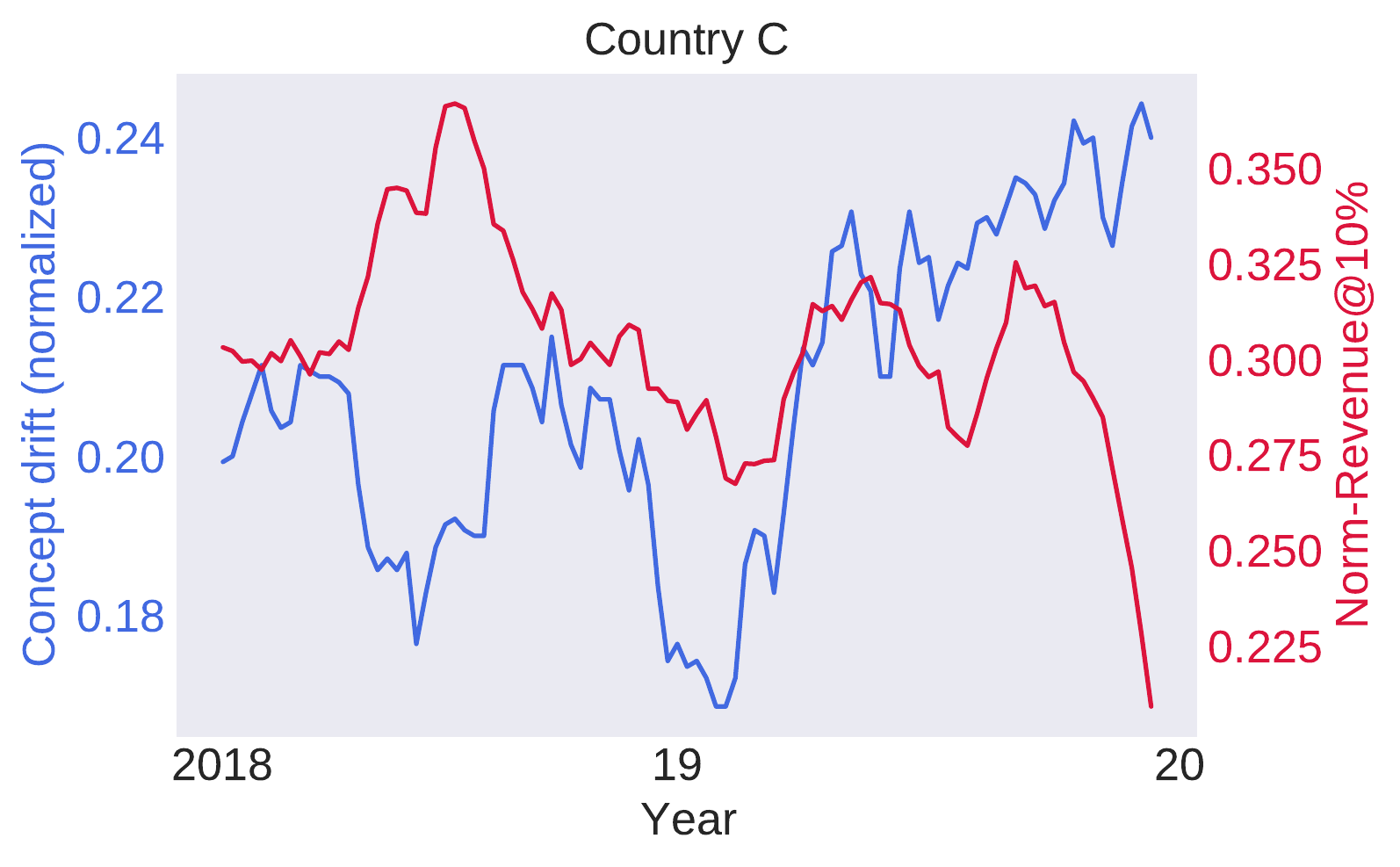}
    \end{subfigure}
     \caption{A strong negative correlation is observed between concept drift in the data (blue line) and performance of full exploitation model (red line). 
     }\label{fig:cd}
     \hfill
   \end{minipage}

\end{figure*}

We analyze the concept drift trends of each country. The EMD-based concept drift score is shown in Figure~\ref{fig:cd-all}. As evidenced from our analysis of Q1, the amount of concept drift seems to be higher in country T and country C, where injecting exploration boosts the performance compared with full exploitation. In general, our EMD-based concept drift score reflects this trend, with the minimal concept drift  scores for country M hovering around 0.15, while the score for T hovers around 0.3. For country C, the full exploitation strategy performs the best before 2018, then the model with 0.2 exploration takes the lead. This suggests that the concept drift starts becoming substantial in 2018. Our score also captures this, with concept drift abruptly soars from around 0.05 before 2018 and plateaus at around 0.2. The EMD-based score for country N peaks at some points then decreases to 0.2.

We conduct further analysis to show the correlation of concept drift score with the model performance. The gist of this analysis is that the performance of the fully-exploitation model will decrease if concept drift occurs. We investigate two datasets with concept drift: T and C (after 2018), and the results are reported in Fig.~\ref{fig:cd}. Pearson correlation test reveals a strong negative correlation between the two trends (correlation coefficient for T: $-0.55$, for C: $-0.23$) with very high confidence (p-value for country T: $1.82 \times{10^{-20}} $, for C: $0.024$). This indicates concept drift indeed harms the typical fraud detection model run by exploitation mechanism. 

\subsubsection{Effectiveness of adaptive strategies in adapting novelties (Q4)}

\begin{figure}[t!]
    \centering
    \begin{subfigure}[b]{0.80\linewidth}
        \centering\captionsetup{width=.95\linewidth}%
        \includegraphics[width=\linewidth]{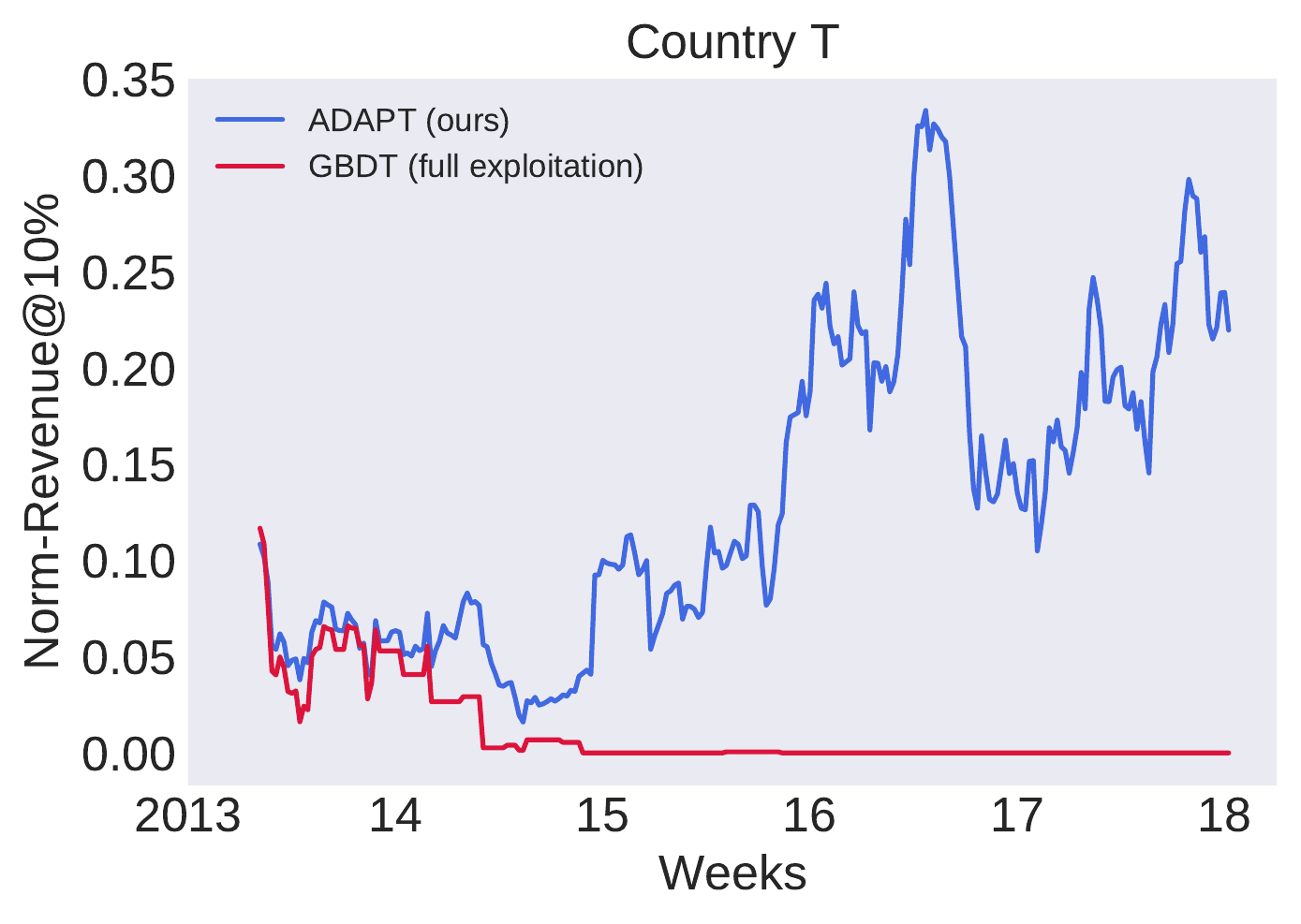}
        \caption{In country \textsf{T}, \ours{} completely dominates full exploitation strategy. Full exploitation completely fails to pick up illicit trades by new importers.}
    \end{subfigure}
    \begin{subfigure}[b]{0.80\linewidth}
        \centering\captionsetup{width=.95\linewidth}%
        \includegraphics[width=\linewidth]{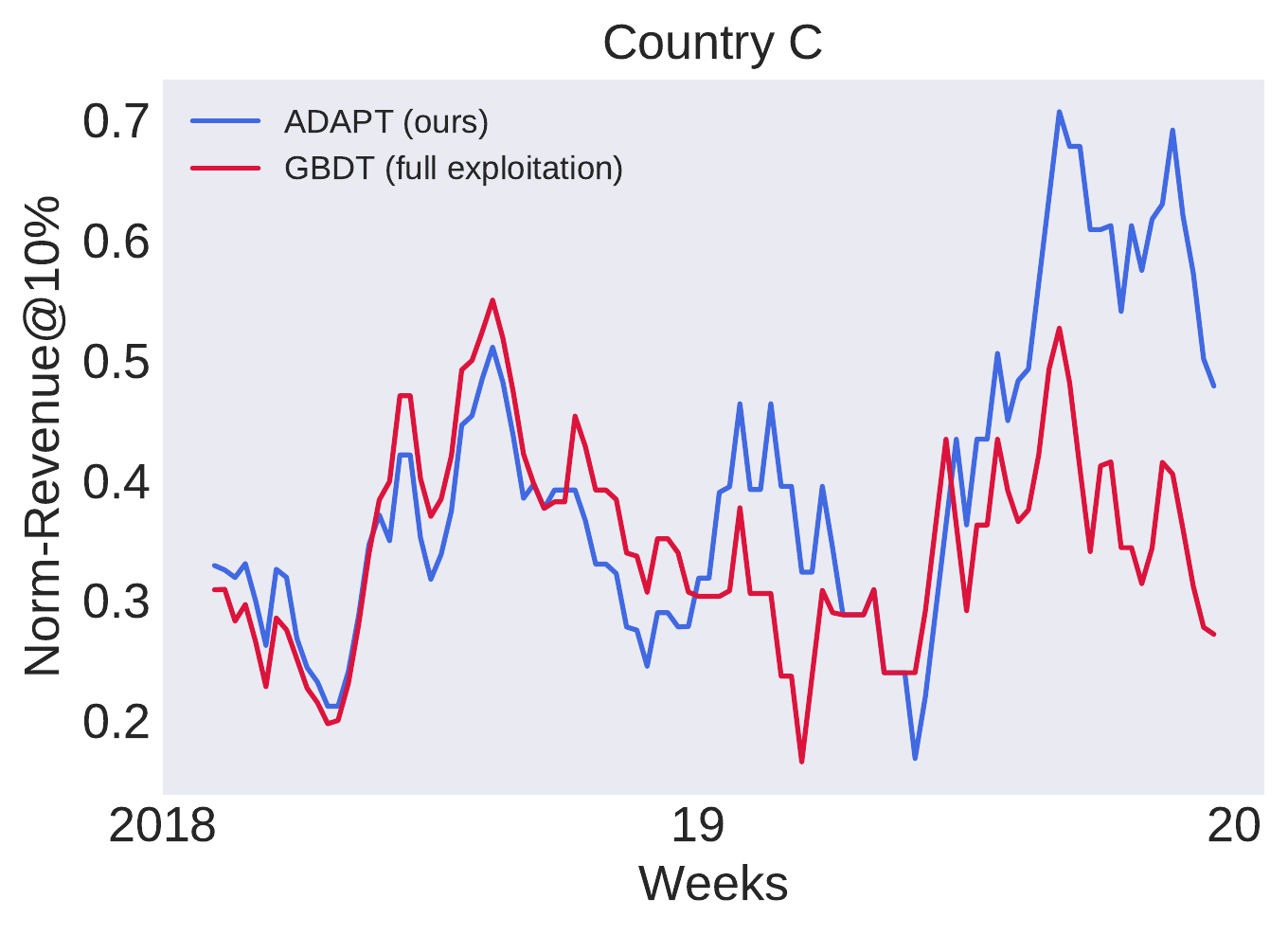}
        \caption{In country \textsf{C}, during duration between 2018 and 2019, two models perform approximately the same. Starting from 2019, \ours{} has risen above GBDT.}
    \end{subfigure}
    \caption{Performance of two strategies on the set of items declared by new importers.}
    \label{fig:novel}
\end{figure}

To evaluate the model's ability to discover new fraud patterns via exploration, we measure the performance on a subset of items declared by new importers. 
We investigate two countries having concept drift and report the results in Fig.~\ref{fig:novel}. 

\ours{} outperforms fully exploitation in both cases. In country C after 2018, \ours{} secure 39.5\% of the revenue on average, compared with 34.6\% secured by full exploitation. In country T, \ours{} captures 13.2\% of possible revenue, significantly outperform full exploitation which secured merely 1.1\% of revenue. In some periods, full exploitation completely fails to pick up any fraud from new importers, while \ours{} still performs robustly. This result supports that \ours{} has explored and inspected the trades declared by new importers, and reused the data successfully to update the model. 

\subsubsection{Ablation study (Q5)}


\begin{table*}[t!]
\centering
\begin{tabular}{c|c|c c c c|c c c c}
\toprule
\multicolumn{2}{l|}{}   & \multicolumn{4}{c|}{\textbf{Average of Norm-Precision}}  & \multicolumn{4}{c}{\textbf{Average of Norm-Revenue}}    \\ \midrule
Country& \multicolumn{1}{c|}{Method} & \begin{tabular}[c]{@{}c@{}}All\\ time\end{tabular} & \begin{tabular}[c]{@{}c@{}}Last 2\\ years\end{tabular} & \begin{tabular}[c]{@{}c@{}}Last 1\\ year\end{tabular} & \begin{tabular}[c]{@{}c@{}}Last 0.5\\ year\end{tabular} & \begin{tabular}[c]{@{}c@{}}All\\ time\end{tabular} & \begin{tabular}[c]{@{}c@{}}Last 2\\ years\end{tabular} & \begin{tabular}[c]{@{}c@{}}Last 1\\ year\end{tabular} & \begin{tabular}[c]{@{}c@{}}Last 0.5\\ year\end{tabular} \\ \midrule
\multirow{6}{*}{C} 
        & \textbf{ADAPT}  & \textbf{0.2803}  & \textbf{0.3106}  & \textbf{0.2985} & \textbf{0.2979} & \textbf{0.2829}  & \textbf{0.3192}  & \textbf{0.3348} & \textbf{0.3660} \\ 
        & \textbf{ADA}  & 0.2785  & 0.3061  & 0.2956 & 0.2851 & 0.2768  & 0.3087  & 0.3336 & 0.3556 \\ 
        & \textbf{APT}  & 0.1806  & 0.1931  & 0.1884 & 0.1799 & 0.1937  & 0.2119  & 0.2386 & 0.2439 \\ 
        \cmidrule{2-10} 

      & \textit{Full exploration}   & 0.1002  & 0.1007  & 0.1020 & 0.1022 & 0.0996  & 0.1030  & 0.1034 & 0.1062 \\  
      & \textit{Full exploitation}  & 0.2776  & 0.2967  & 0.2803 & 0.2636 & 0.2613  & 0.2707  & 0.2736 & 0.2875 \\ 
      & \textit{Oracle} & 0.2661  & 0.3164  & 0.3129 & 0.2992 & 0.2776  & 0.3343  & 0.3548 & 0.3767 \\ 
      \midrule
\multirow{6}{*}{T} 
        & \textbf{ADAPT}  & \textbf{0.3520}  & \textbf{0.3827}  & \textbf{0.4154} & 0.3906 & \textbf{0.4385}  & \textbf{0.4848}  & \textbf{0.5442} & 0.5651   \\ 
        & \textbf{ADA}  & 0.3502  & 0.3794  & 0.4127 & \textbf{0.3927} & 0.4345  & 0.4801  & 0.5435 & \textbf{0.5713}   \\ 
        & \textbf{APT}  & 0.3145  & 0.3214  & 0.3589 & 0.3284 & 0.3934  & 0.4072  & 0.4705 & 0.4692   \\ 
        
        \cmidrule{2-10} 
      & \textit{Full exploration}   & 0.1023  & 0.1035  & 0.1049 & 0.0998 & 0.0987  & 0.0974  & 0.0967 & 0.0977 \\  
      & \textit{Full exploitation}  & 0.1442  & 0.1108  & 0.1018 & 0.0962 & 0.1405  & 0.0895  & 0.0899 & 0.1094 \\ 
      & \textit{Oracle} & 0.3535  & 0.4243  & 0.4687 & 0.4466 & 0.4386  & 0.5424  & 0.6131 & 0.6437    
      \\ \midrule
\multirow{6}{*}{M} 
        & \textbf{ADAPT}  & \textbf{0.5873}  & \textbf{0.7721 } & \textbf{0.8221} & \textbf{0.8703} & \textbf{0.5615}  & \textbf{0.7304}  & \textbf{0.8087} & \textbf{0.8859} \\ 
        & \textbf{ADA}  & 0.5001  & 0.6638  & 0.6977 & 0.7557 & 0.4704  & 0.6222  & 0.6860 & 0.7439 \\ 
        & \textbf{APT}  & 0.3804  & 0.5554  & 0.6413 & 0.8017 & 0.3469  & 0.5147  & 0.6651 & 0.8382 \\ 
        \cmidrule{2-10} 
      & \textit{Full exploration}   & 0.0204  & 0.0218  & 0.0203 & 0.0225  & 0.0168  & 0.0167  & 0.0167 & 0.0178  \\ 
      & {\textit{Full exploitation (Oracle)}}  & {0.6549}     & {0.8936}     & {0.9457}    & {0.9642} & {0.6189}     & {0.8426}     & {0.9289}    & {0.9664} \\ \midrule
\multirow{6}{*}{N} 
        & \textbf{ADAPT}  & \textbf{0.7658}  & \textbf{0.7953}  & \textbf{0.8435} & 0.8567 & \textbf{0.7099}  & \textbf{0.7355}  & \textbf{0.7408} & \textbf{0.7500} \\ 
        & \textbf{ADA}  & 0.7656  & 0.7953  & 0.8431 & \textbf{0.8570} & 0.7079  & 0.7317  & 0.7345 & 0.7388 \\ 
        & \textbf{APT}  & 0.5208  & 0.5098 & 0.5348 & 0.5100 & 0.4787  & 0.4894  & 0.4954 & 0.4644 \\ 
        \cmidrule{2-10} 
      & \textit{Full exploration}   & 0.1008  & 0.1015  & 0.1008 & 0.1001 & 0.0999  & 0.0972  & 0.0912 & 0.0871 \\  
      & \textit{Full exploitation (Oracle)}  & {0.7947}     & {0.8271}     & {0.8648}    & {0.8738}    & {0.7396}     & {0.7604}     & {0.7650}    & {0.7757}     
 \\ \bottomrule
\end{tabular}
\caption{Precision and revenue of our proposed methods and baselines. Performance over the last $k$ years are averaged and reported.}
\label{tbl:main}
\end{table*}

To validate the contribution of each component of \ours{}, we conduct an ablation study by examining the performance after removing each component. Precision trends between these three methods are shown in Fig.~\ref{fig:rada} and the averaged results are in Table.~\ref{tbl:main}. 

\begin{itemize}
    \item \textbf{\ours{}}: Use both performance signals and concept drift signals.
    \item \textbf{w/o concept drift signal (APT)}: EXP3.S-based method without filtering (ignore input from concept drift scorer)
    \item \textbf{w/o performance signal (ADA)}: Directly use the concept drift score as the exploration rate (ignore performance signal)
\end{itemize}

\begin{figure*}[h!]
    \centering
    \begin{subfigure}[b]{.38\linewidth}
        \centering\captionsetup{width=.95\linewidth}%
\includegraphics[trim={0 0 5.5cm 0},clip,width=\linewidth]{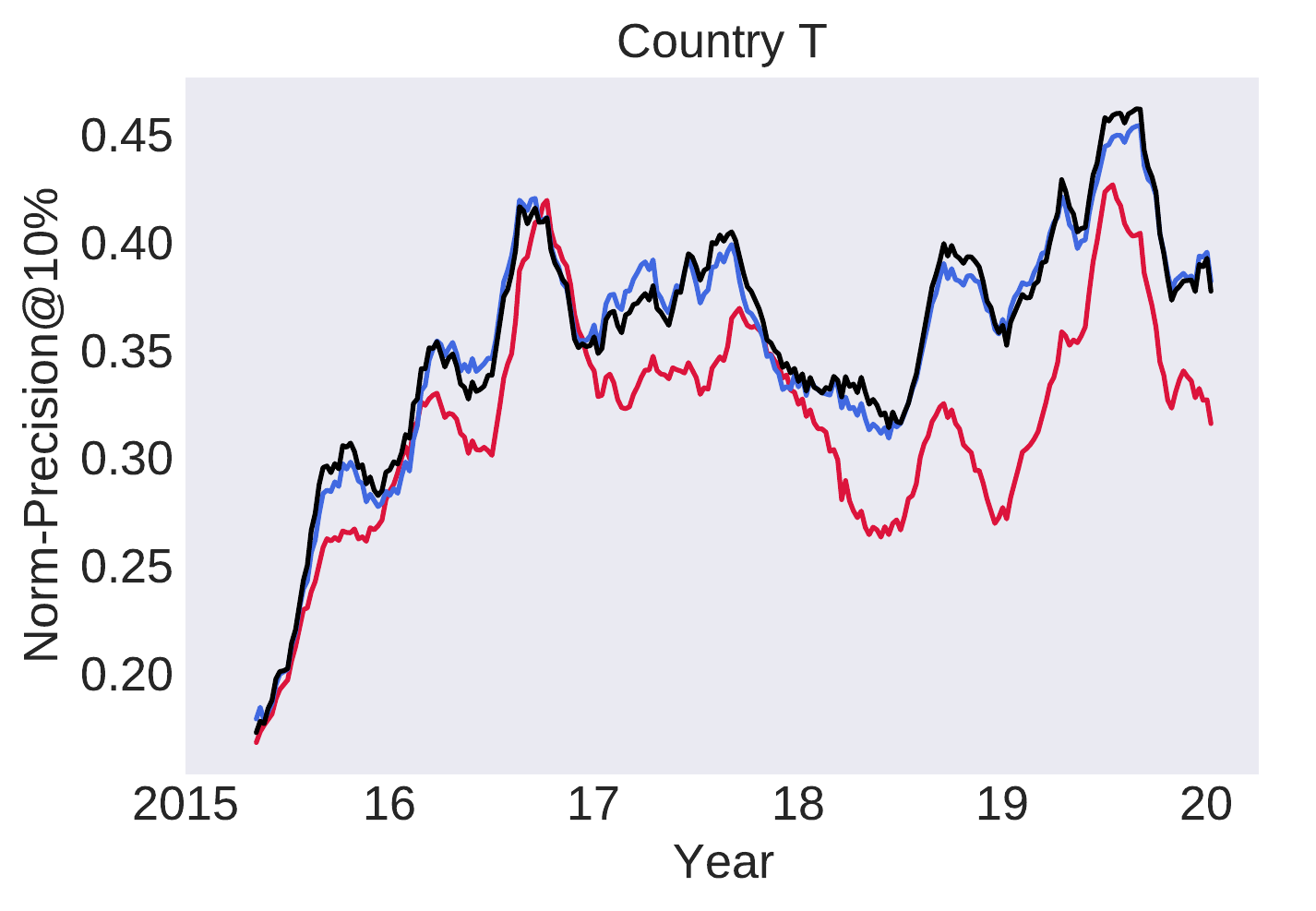}
    \end{subfigure}
    \begin{subfigure}[b]{.38\linewidth}
        \centering\captionsetup{width=.95\linewidth}%
        \includegraphics[trim={0 0 5.5cm 0},clip,width=\linewidth]{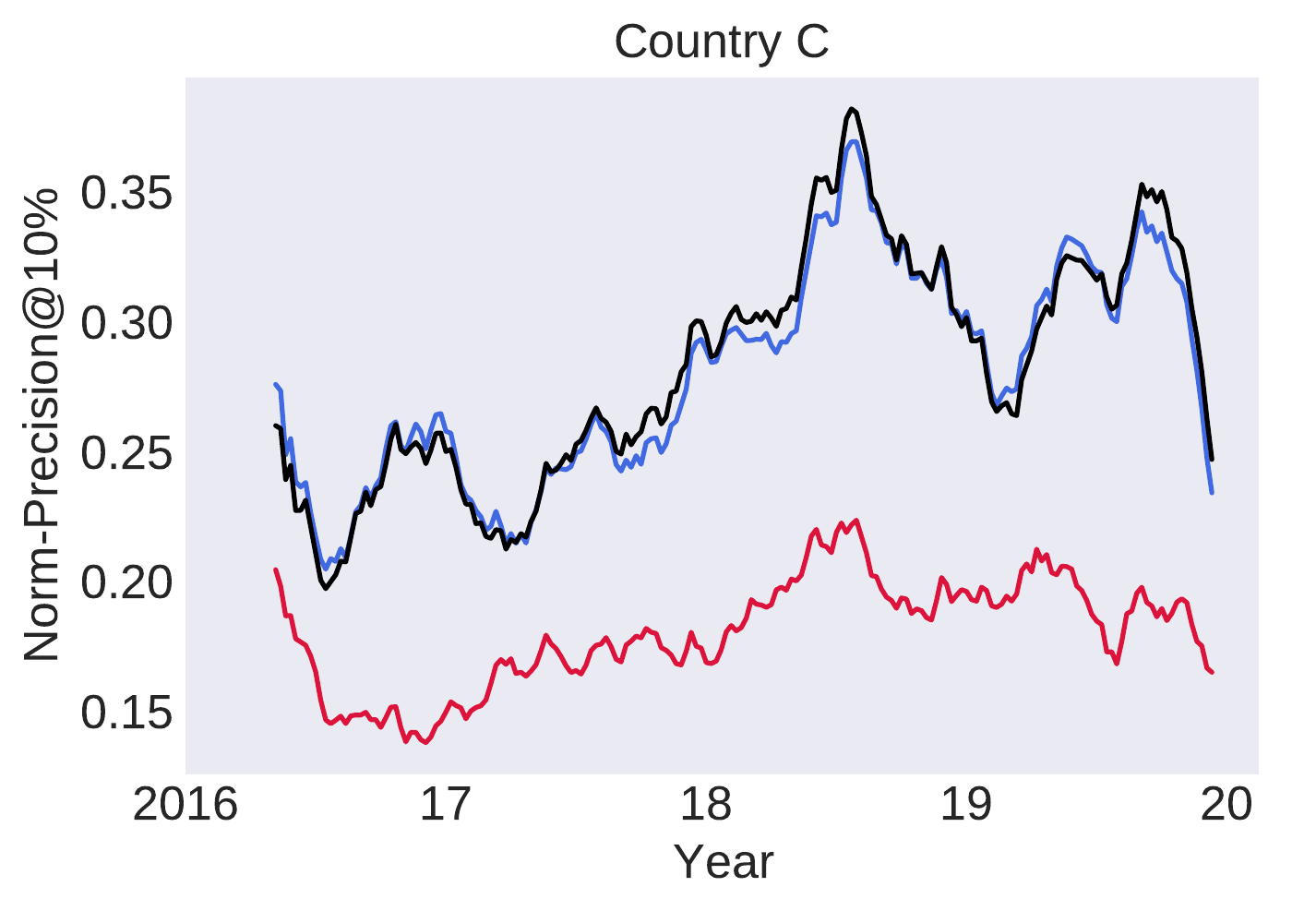}
    \end{subfigure}
    \begin{subfigure}[b]{.38\linewidth}
        \centering\captionsetup{width=.95\linewidth}%
\includegraphics[trim={0 0 5.5cm 0},clip,width=\linewidth]{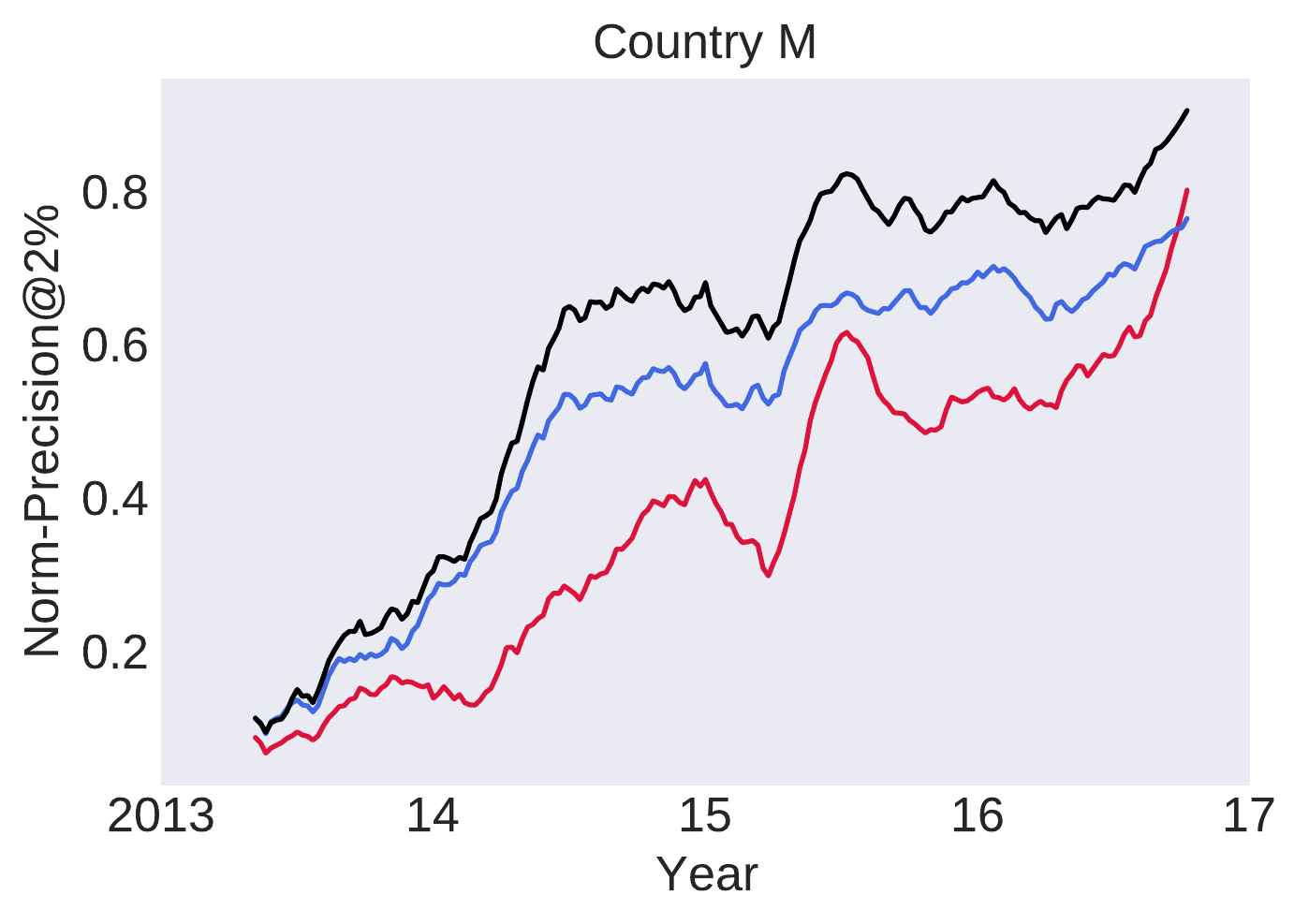}
    \end{subfigure}
    \begin{subfigure}[b]{.38\linewidth}
        \centering\captionsetup{width=.95\linewidth}%
        \includegraphics[trim={0 0 5.5cm 0},clip,width=\linewidth]{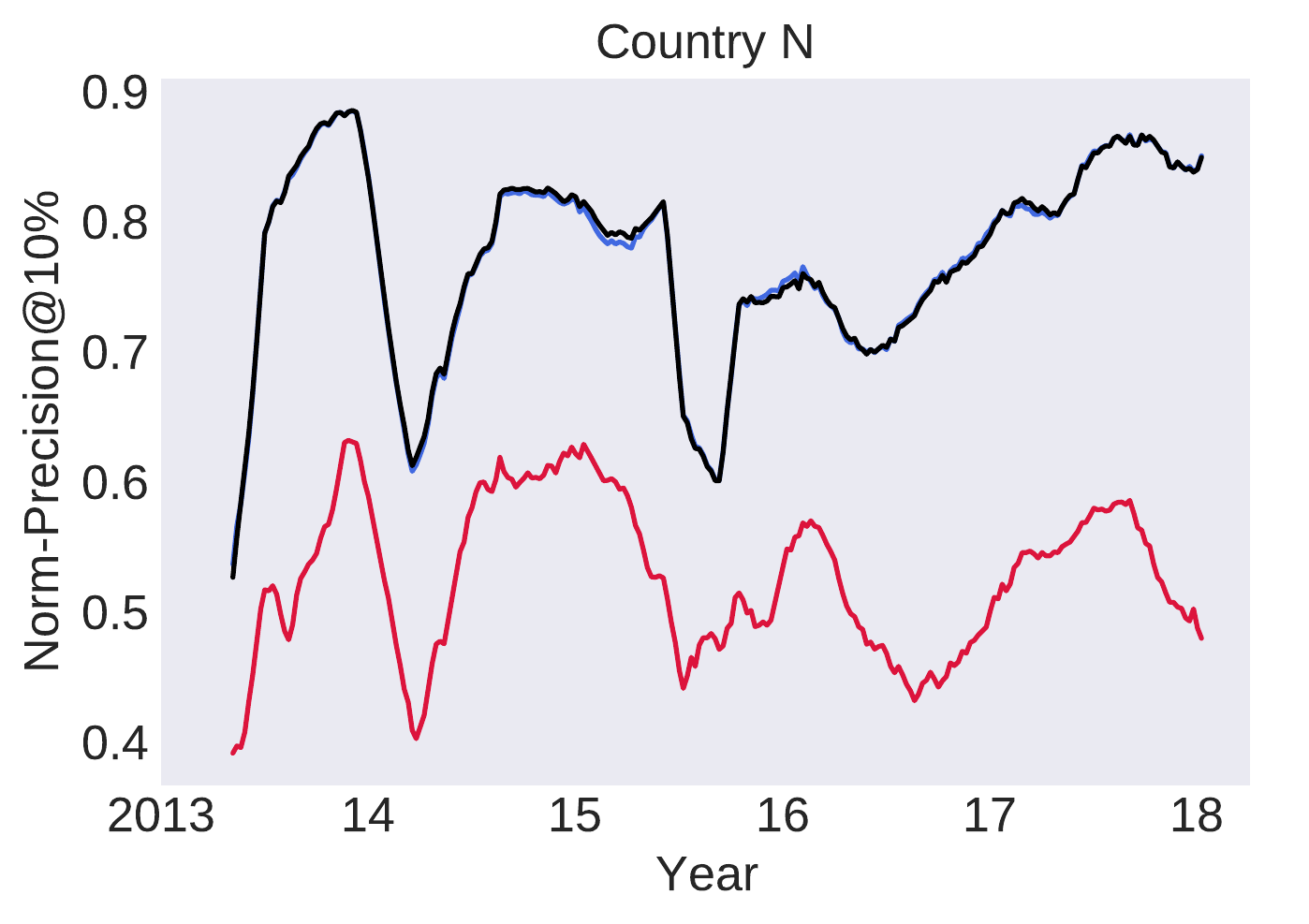}
    \end{subfigure}

    \begin{subfigure}[b]{0.20\linewidth}
        \centering
        \includegraphics[trim={15cm 7.5cm 0 1cm},clip,width=\linewidth]{figures/ablation-real-c-norm-precision.pdf}
    \end{subfigure}
    \caption{\textbf{Performance of \ours{} with its variants APT and ADA}. \ours{} outperforms APT in all countries and showed similar or higher performance than ADA.}
    \label{fig:ablation}
\end{figure*}

In all four countries, the performance of \ours{} is better than APT and closest to the best hybrid strategy, indicating that the concept drift successfully guides the APT arm selector to concentrate on the correct range of ratios.

In country M and country C, precision and revenue of \ours{} are better than ADA. In the other two countries, they perform very similarly. This suggests that the MAB-based model with the performance signal could improve upon ADA by considering neighboring ratios with a history of good performance. In cases with concept drift or not, \ours{} is expected to be more robust than ADA.




\section{Conclusion}

This paper examines the importance of the exploitation-exploration ratio and proposes an algorithm, namely \ours{}, that combines two signals for dynamically deciding this value. By conducting experiments in online settings with four different countries' customs data, we show that \ours{} can perform commensurately with the oracle. Moreover, since the proposed algorithm is based on the shifts in data distribution and the preceding model performance, it does not require extensive tuning and careful result observation over a long period.
We also show the importance of concept drift detection in avoiding model failures over time when changes in data distribution occur, which are frequent in customs trades. In the customs setting, the potential concept drift, indicated by \ours{}, can alert customs officers to new fraud patterns that require more cautious inspection. We will open-source the code so that \ours{} can be employed by the customs administrations to ease and improve the trade inspection process.

\section*{Acknowledgment}
This work was supported by the Institute for Basic Science (IBS-R029-C2, IBS-R029-Y4). We thank World Customs Organization and their partner countries to support their datasets. 
\balance
\bibliographystyle{IEEEtran}
\bibliography{main}


\end{document}